\documentclass[11pt, a4paper, twocolumn, copyright, goog]{google}

\usepackage[numbers, sort&compress, round]{natbib}
\usepackage[utf8]{inputenc} % allow utf-8 input
\usepackage[T1]{fontenc}    % use 8-bit T1 fonts
\usepackage{hyperref}       % hyperlinks
\usepackage{url}            % simple URL typesetting
\usepackage{booktabs}       % professional-quality tables
\usepackage{amsfonts}       % blackboard math symbols
\usepackage{nicefrac}       % compact symbols for 1/2, etc.
\usepackage{microtype}      % microtypography
\usepackage{xcolor}         % colors
\usepackage{graphicx}
\usepackage{subcaption}
\usepackage{multirow}       % Required for the \multirow command in the Summary Table
\usepackage{makecell}       % Required for the \makecell command to stack text/retention rates in the Detailed Table
\usepackage{xspace}
\usepackage{amsmath}
\usepackage{tabularx}
\usepackage{array}
\usepackage{bbm}
\definecolor{plotblue}{HTML}{1F77B4}
\definecolor{plotred}{HTML}{D62728}
\newcommand{\ours}{PARCEL\xspace}

\uselogo{} 

\title{PARCEL: Pool-Anchored Resampling with Conditioned Elastic Queries for Efficient Vision-Language Understanding}

\correspondingauthor{skuzucu@mpi-inf.mpg.de, ferjad@google.com}

\renewcommand{\today}{June 2026}

\author[2, $\dagger$]{Selim Kuzucu}
\author[1]{Alessio Tonioni}
\author[1]{Vasile Lup}
\author[1]{Bernt Schiele}
\author[1, 3]{Federico Tombari}
\author[1]{Muhammad Ferjad Naeem}

\affil[1]{\thepa{}{}}
\affil[2]{Max Planck Institute for Informatics, SIC}
\affil[3]{Technical University of Munich}
\affil[$\dagger$]{Work done while interning at Google.}

\begin{abstract}
Large Vision-Language Models (LVLMs) map visual inputs into dense token sequences, imposing a quadratic computational bottleneck for inference. Elastic visual-token compression addresses this by training a single model that can run at multiple visual-token budgets. However, existing approaches struggle under aggressive compression. Spatial-only compression, as in nested pooling, behaves as an imperfect low-pass filter and induces spectral aliasing that obscures fine-grained detail. Query-only compression, as in nested query resampling, replaces explicit grid-aligned tokens with non-local summaries and substantially degrades spatial grounding. To resolve this representational conflict, we introduce PARCEL
(\textbf{P}ool-\textbf{A}nchored \textbf{R}esampling with \textbf{C}onditioned \textbf{EL}astic Queries for Efficient Vision-Language Understanding),
a visual tokenization architecture that dynamically partitions the labor of feature extraction. PARCEL establishes spatial pool tokens as low-frequency layout anchors and conditions elastic query tokens on these anchors through Pool-Conditioned Query Resampling. This encourages query tokens to focus on complementary visual features rather than redundant spatial mapping. Extensive evaluations across 27 benchmarks show that PARCEL improves the performance-efficiency Pareto frontier, consistently outperforming existing matryoshka baselines across visual-token budgets while preserving the ``train once, deploy anywhere'' paradigm.
\end{abstract}

\begin{document}

\maketitle

\section{Introduction}

Large Vision-Language Models (LVLMs) \cite{SigLIP, SigLIP2, QwenVL, InternVL2_5, InternVL3, PG2, PG} have achieved remarkable success across a wide range of multimodal tasks, spanning video understanding, dense recognition, and generic visual question answering.
Despite this success, LVLMs face an input-side bottleneck: images or videos are often represented with hundreds or thousands of visual tokens before being processed by the language decoder.
This directly increases the sequence length of the Transformer \cite{Vaswani}, whose self-attention cost scales quadratically with the number of tokens.
Prior LVLMs \cite{PG2, LLaVA-OV, LLaVA-1_5} show that increasing the visual-token budget often improves visual representation quality and downstream performance, but this comes at a steep compute and memory cost, hindering ubiquitous deployment.
%%%%%%%%%%%%%%%%%%%%%%%%%%%%%
\begin{figure*}[t]
    \centering
    \begin{minipage}[c]{0.45\textwidth}
        \centering
        \includegraphics[width=\linewidth]{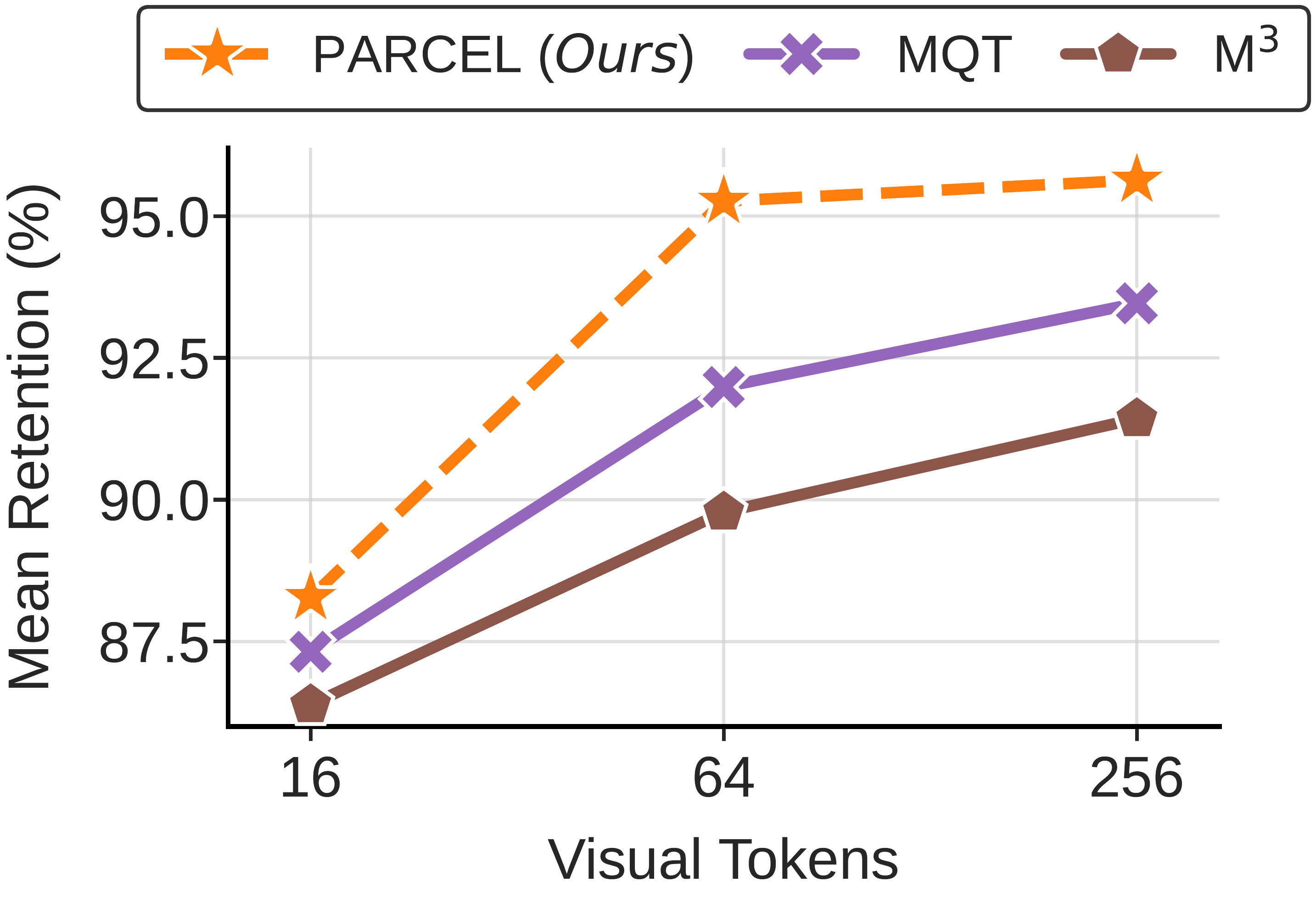}
    \end{minipage}
    \hspace{0.04\textwidth}
    \begin{minipage}[c]{0.42\textwidth}
        \centering
        \small
        \renewcommand{\arraystretch}{1.08}
        \setlength{\tabcolsep}{3.5pt}
        \begin{tabular}{ccccc}
    \toprule
    \textbf{Budget} & \makecell{\textbf{Image}\\\textbf{TFLOPs}} & \makecell{\textbf{Video}\\\textbf{TFLOPs}} & \makecell{\textbf{Image}\\\textbf{KV}} & \makecell{\textbf{Video}\\\textbf{KV}} \\
    \midrule
    16  & 1.0T & 4.9T  & 15MB & 33MB \\
    64  & 1.2T & 8.2T  & 20MB & 111MB \\
    256 & 2.0T & 24.3T & 39MB & 423MB \\
    \bottomrule
\end{tabular}
    \end{minipage}
\caption{\textbf{Aggregate retention--efficiency trade-off.}
\textit{Left}: mean retention relative to Vanilla PG2 over 27 benchmarks and 3 seeds.
\textit{Right}: theoretical \ours prefill FLOPs and LLM KV-cache costs by visual-token budget. KV-cache is identical across methods at matched budgets; FLOP differences are small since shared ViT/LLM terms dominate and only connector overhead differs by a small margin. \ours outperforms MQT and M$^3$ at matched budgets; lower visual-token budgets reduce compute and KV-cache costs versus uncompressed PG2 for image and 16-frame video.}
    \label{fig:retention_flops_memory}
\end{figure*}

To mitigate this computational burden, prior works explore static visual token compression techniques, including dropping \cite{FastV, PyramidDrop, VScan, MustDrop, HiRED, DivPrune, PruneVid}, merging \cite{ToFu, LLaVAPruMerge}, and projection \cite{Honeybee, NVILA, LLaVA-SP, TokenPacker}.
While effective at reducing inference costs, these approaches typically produce fixed-length visual representations, forcing practitioners to choose a single operating point before deployment.
This creates a strict trade-off: an efficiency-optimized model permanently sacrifices fine-grained visual detail, whereas a high-resolution model remains computationally prohibitive for lightweight and latency-sensitive applications.
In practice, available resources vary across devices and latency targets, especially when accommodating diverse input domains from images to videos.
Elastic inference, a ``train once, deploy anywhere'' approach that supports multiple budgets after a single stage of training, has therefore emerged as a valuable and practical deployment goal \cite{MRL, MatFormer, M3, MQT, AIM, ATP-LLaVA, ATP}.

To achieve this elasticity, recent advances adapt Matryoshka-style representation learning \cite{MRL} to LVLM visual tokenization.
These efforts primarily branch into two distinct architectural paradigms: rigid spatial downsampling \cite{M3} and non-local query resampling \cite{MQT}.
Matryoshka Multimodal Models (M$^3$) \cite{M3} represent the former, constructing a nested token structure through successive multi-scale spatial average pooling.
Conversely, the Matryoshka Query Transformer (MQT) \cite{MQT} achieves elasticity using a query transformer paired with a nested dropout strategy \cite{NestedDropout, Dropout}.
While both successfully establish an elastic inference paradigm, they introduce opposing representational bottlenecks at highly constrained token budgets.
As we formally analyze in Section~\ref{sec:methodology}, the rigid spatial downsampling in M$^3$ acts as an imperfect low-pass filter.
This induces spectral aliasing that blurs the high-frequency semantic details required for resolution-sensitive tasks, such as chart reasoning and text-centric visual-question answering.
In contrast, query resampling employed by MQT sacrifices explicit spatial relationships in favor of non-local learned summaries, reducing spatial grounding and dense localization capabilities.

\begin{figure*}[ht]
    \centering
    \newlength{\teaserheight}
    \setlength{\teaserheight}{0.17\textheight}

    \begin{subfigure}[t]{0.25\textwidth}
        \centering
        \includegraphics[height=\teaserheight]{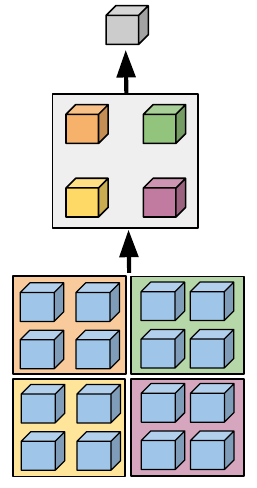}
        \caption{M$^3$ \cite{M3}}
        \label{fig:teaser_m3}
    \end{subfigure}
    \hspace{0.035\textwidth}
    \begin{subfigure}[t]{0.28\textwidth}
        \centering
        \includegraphics[height=\teaserheight]{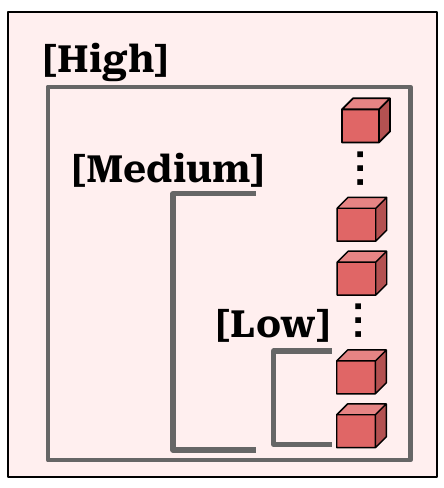}
        \caption{MQT \cite{MQT}}
        \label{fig:teaser_mqt}
    \end{subfigure}
    \hspace{0.035\textwidth}
    \begin{subfigure}[t]{0.34\textwidth}
        \centering
        \includegraphics[height=\teaserheight]{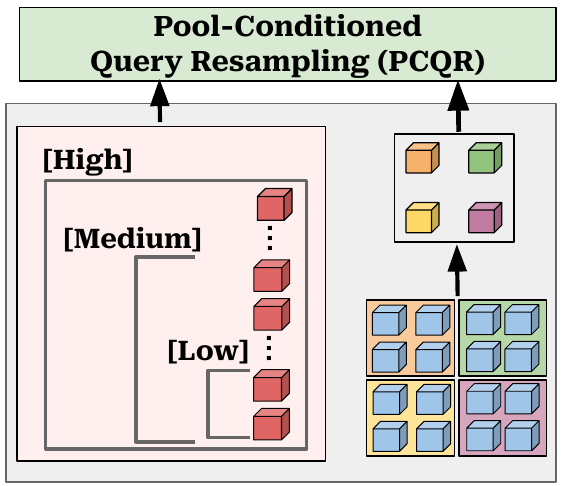}
        \caption{\ours \textit{(Ours)}}
        \label{fig:teaser_parcel}
    \end{subfigure}

    \caption{\textbf{High-Level Overview of MQT, M$^3$ and \ours~(Ours).}
    M$^3$ compresses visual features through rigid spatial pooling, MQT uses elastic query tokens, and \ours combines spatial anchor tokens with pool-conditioned query resampling, allowing it to compress more effectively.}
    \label{fig:method_high_level_teaser}
\end{figure*}
%%%%%%%%%%%%%%%%%%%%%%%%%%%%%%
To resolve these representational conflicts, we propose \textbf{\ours} 
(\textbf{P}ool-\textbf{A}nchored \textbf{R}esampling with \textbf{C}onditioned \textbf{EL}astic Queries for Efficient Vision-Language Understanding)
, a visual tokenization architecture that dynamically partitions the labor of feature extraction as shown in Figure~\ref{fig:method_high_level_teaser}.
\ours encourages a division of labor between spatial anchors directly coming from the vision transformer and learnable query tokens. The spatial pool tokens anchor the low-frequency geometric layout.
We then introduce a supporting set of nested-dropout query tokens that are explicitly conditioned on these spatial anchors.
Operating as dedicated ``semantic explorers,'' these pool-aware queries recover the complementary visual signal that standard pooling discards.
Our contributions are as follows:
\begin{itemize}[leftmargin=0.5cm, itemsep=0.05cm, topsep=0.1cm]
    \item \textbf{Analysis of Spectral Bottlenecks:} We formalize and empirically demonstrate the opposing representational bottlenecks present in current elastic LVLMs.
    Specifically, we show that rigid spatial average pooling in M$^3$ exhibits spectral signatures akin to aliasing, while non-local query-based resampling in MQT substantially degrades dense spatial understanding under compression.
    
    \item \textbf{The \ours Architecture:} To resolve the above bottlenecks, we introduce a hybrid visual connector that dynamically partitions the labor of feature extraction.
    By means of our \textbf{Pool-Conditioned Query Resampling} mechanism, we combine the geometric stability of rigid spatial anchors with the high-frequency expressivity of dedicated semantic explorer queries.
    
    \item \textbf{Budget-Aware Routing and Pareto Efficiency:} We design a dynamic routing strategy that enables a single model to seamlessly operate across varying inference budgets (from 16 to 256 tokens). This approach improves the performance-efficiency Pareto frontier across 27 diverse vision-language benchmarks, spanning video understanding, dense recognition, and VQA.
\end{itemize}

\section{Background and Related Work}
\label{ref:related-works}
\textbf{Visual Token Compression in LVLMs.} The quadratic complexity of attention has driven research into reducing LVLM visual tokens via dropping \cite{FastV, PyramidDrop, VScan, MustDrop, HiRED, DivPrune, PruneVid, EAdaPrune, VLM-Pruner, VisionMeetsLanguage}, merging \cite{ToFu, LLaVAPruMerge}, spatial reshaping and resolution adaptation \cite{InternVL2_5, InternVL3, ResAdapt, VisualContextCompressor}, projecting \cite{Honeybee, NVILA, LLaVA-SP, TokenPacker, DeltaLLaVA}, and query-based resampling \cite{QwenVL, QueCC}.
As highlighted by recent surveys on efficient multimodal learning \cite{wang2025models}, optimizing these architectural bottlenecks remains an active research frontier.
While several recent methods \cite{FEATHER, NUWA, TangoTamer, LaPrune, AreWeSolvingRight, AGILEPRUNER, DualComp} address positional biases and spatial distortions caused by aggressive dropping, they mostly remain training-free, post-hoc optimizations rather than jointly trained elastic connectors.
However, as Kong et al. \cite{TokenReductionBeyond} note, the lack of gradient-aligned learning often limits post-hoc compression methods under dynamic efficiency-performance trade-offs.
Further recent advances explore hybrid and distillation-based compression, \textit{e.g.}, using visual bottleneck or summary tokens \cite{VoCo, Adaptive-VoCo, HTC-VLM, Fwd2Bot}.
Unlike our approach, these methods operate predominantly inside the LLM.
Because they bottleneck text-to-visual attention or rely on static quantization rather than elastic connector-level tokenization, they do not provide native inference-time elasticity within a unified visual connector.

\textbf{Matryoshka Representation Learning.} To achieve inherent deployment elasticity without retraining, Matryoshka Representation Learning (MRL) \cite{MRL} encodes information at multiple granularities within a single nested structure.
This ``train once, deploy anywhere'' paradigm spans diverse architectures, including nested transformers \cite{MatFormer} and Mixture-of-Experts routing \cite{M-MoE, QMoP, MoME}.
Within the generative and broader representation learning domains, hierarchical tokenizers \cite{FlexTok, SEMANTICIST} and adaptive autoencoders like ElasticTok \cite{ElasticTok} utilize nested dropout to resample 2D images into variable-length 1D token sequences, demonstrating that sequence truncation can naturally align with coarse-to-fine generation. 

\textbf{Adaptive and Elastic Visual Tokenization.} Following the matryoshka learning paradigm, recent works aim to equip visual-token compression with the ability to perform inference at multiple visual-token budgets after a single stage of training.
Recent methods \cite{AIM, ATP-LLaVA, ATP} introduce adaptive inference via dynamic, input-dependent token pruning.
Other works realize this flexibility by allowing the user to specify the token count, such as Mask-LLaVA \cite{MaskLLaVA}.
Among these budget-aware architectures, Matryoshka Multimodal Models ($\text{M}^3$) \cite{M3} and the Matryoshka Query Transformer (MQT) \cite{MQT} stand out as representative paradigms that achieve structural elasticity via two distinct mechanisms.
$\text{M}^3$ utilizes multi-scale successive spatial average pooling to obtain elasticity across multiple visual token budgets.
Conversely, MQT employs nested-dropout query resampling \cite{NestedDropout, Dropout}, where the latent query sequence is truncated to randomly sampled lengths during training for elasticity.

While both successfully establish elastic inference, they expose opposing bottlenecks under compression.
The rigid pooling in M$^3$ behaves like a spatial downsampling operator and is prone to spectral aliasing that weakens fine-grained detail.
In contrast, MQT relies on non-local query resampling, which is suboptimal for spatial understanding.
By explicitly dividing the labor of feature extraction, \ours utilizes spatial anchors to retain the geometric layout, allowing the pool-conditioned query tokens to capture complementary high-frequency visual features.
Furthermore, standard VQA tasks widely used in prior work often saturate, offering limited insight into the effects of visual-token compression~\cite{AreWeBenchmarkingRight}.  
Consequently, we validate \ours across a wide range of resolution-sensitive, dense reasoning and video understanding tasks.

\section{\ours: Pool-Anchored Resampling with Conditioned Elastic Queries}
\label{sec:methodology}

\begin{figure*}[t]
    \centering
    % Left Panel: Phase 2 at 64 Tokens
    \begin{subfigure}[b]{0.42\textwidth}
        \centering
        \includegraphics[width=\textwidth]{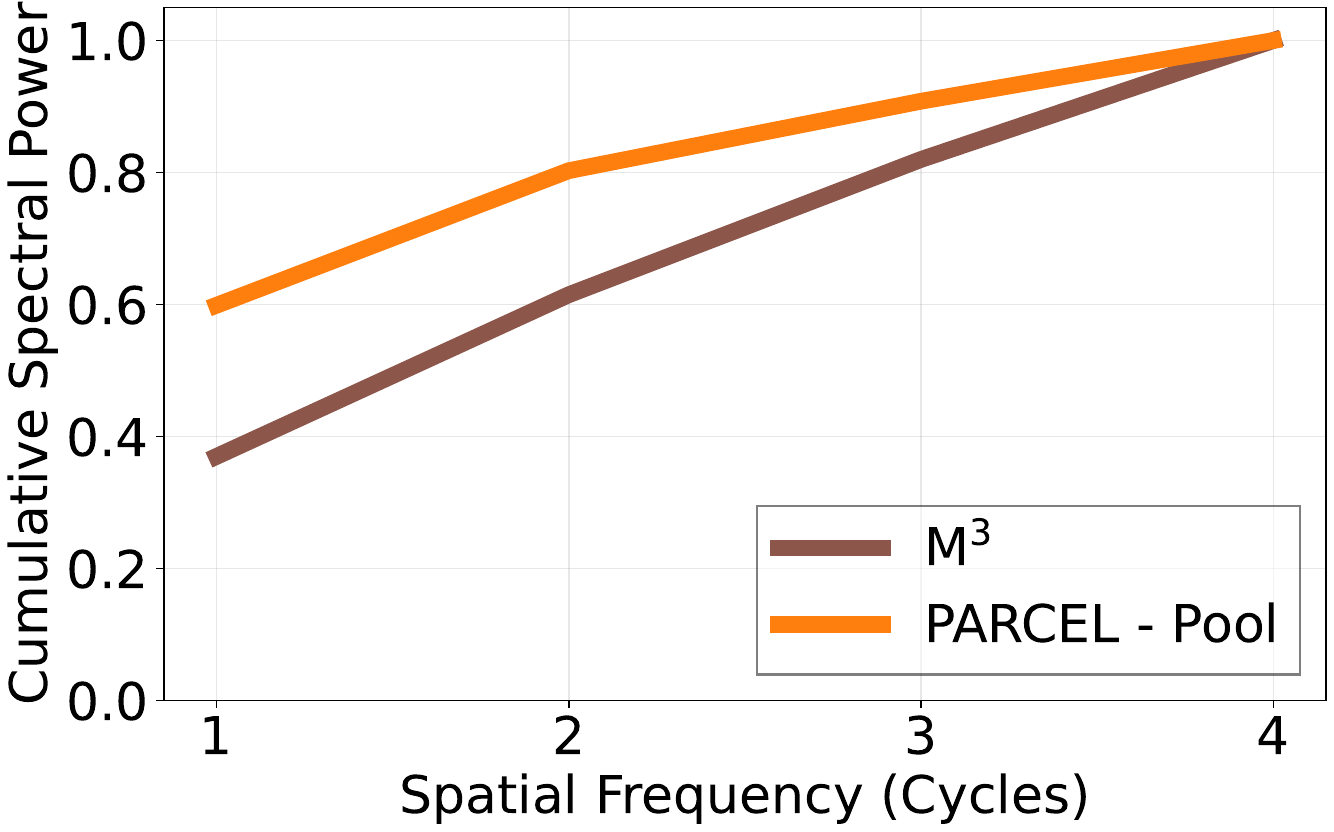}
        \caption{Baseband Concentration (64 Tokens)}
        \label{fig:spectral_a}
    \end{subfigure}
    \hfill
    % Right Panel: Phase 3 at 256 Tokens
    \begin{subfigure}[b]{0.42\textwidth}
        \centering
        \includegraphics[width=\textwidth]{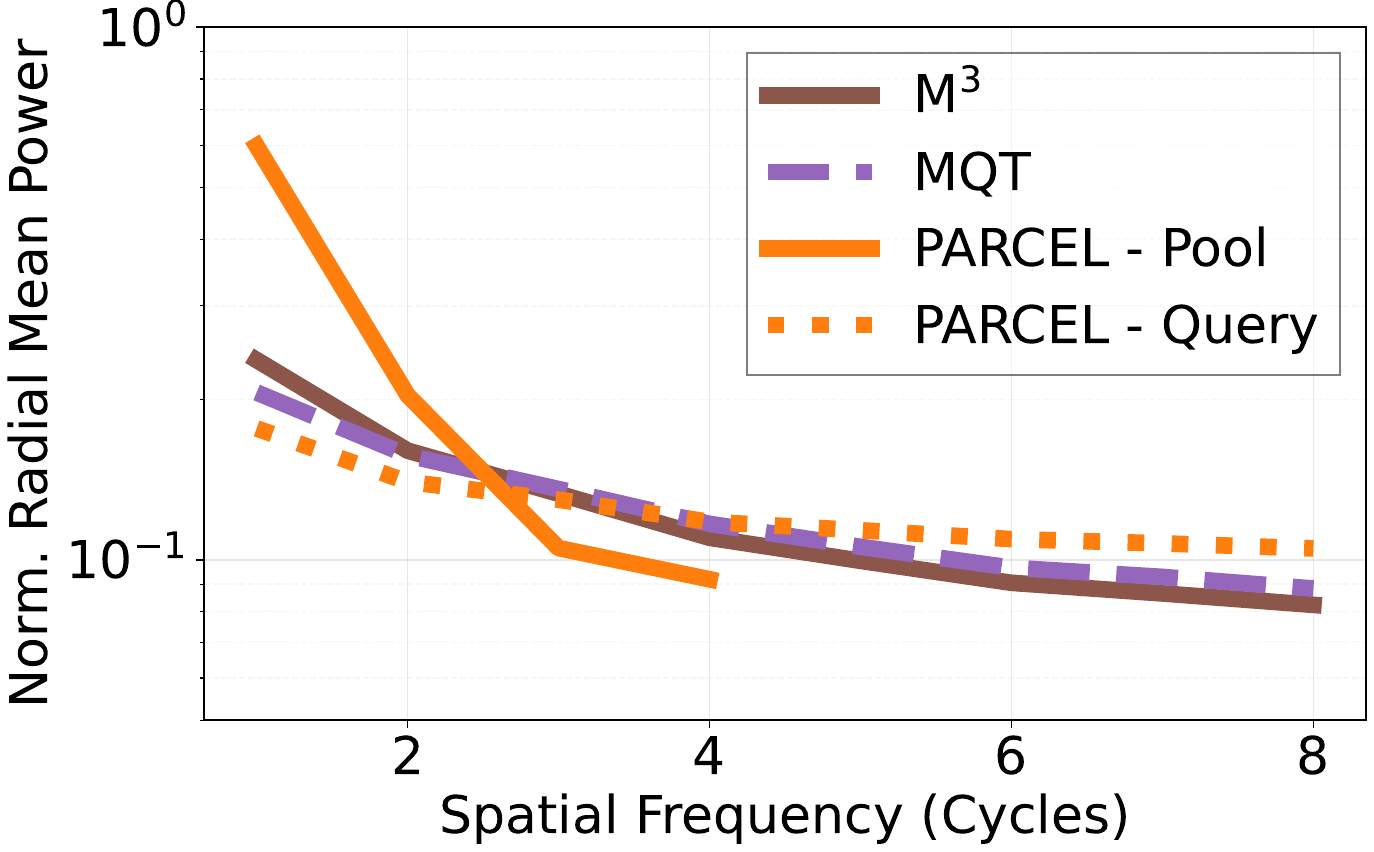}
        \caption{Spectral Disentanglement (256 Tokens)}
        \label{fig:spectral_b}
    \end{subfigure}
    
\caption{\textbf{Spectral decoupling on ChartQA.}
\textbf{(a)} Cumulative spectral power shows that \ours concentrates compressed spatial tokens at low frequencies faster than \(M^3\), indicating a cleaner baseband for spatial anchoring.
\textbf{(b)} Normalized radial mean power shows complementary token roles: pool tokens focus on low-frequency layout, while query-attended ViT feature footprints retain higher-frequency detail beyond the pooled grid.
This separation aligns with ChartQA gains of \(\mathbf{+4.7}\) and \(\mathbf{+3.4}\) points over \(\text{M}^3\) at 64 and 256 tokens, respectively.}
\label{fig:spectral_analysis}
\end{figure*}

While existing matryoshka visual token compression techniques successfully reduce the quadratic cost of visual tokens in LVLMs, relying strictly on either successive spatial average pooling \cite{M3} or non-local query resampling \cite{MQT} degrades the visual representation in complementary ways under aggressive compression.
Rigid spatial downsampling induces spectral leakage that blurs fine-grained visual details, whereas query-only resampling replaces explicit grid-aligned tokens with non-local summaries, weakening the grid-to-token correspondence needed for dense spatial grounding.
To address this representational conflict, we introduce \ours, an architecture designed to dynamically partition the labor of feature extraction.
First, spatial pool tokens serve as 2D anchors for the low-frequency geometric layout.
Second, \textbf{Pool-Conditioned Query Resampling} provides a complementary pathway, where learnable query tokens are explicitly conditioned on the spatial anchors before interacting with the raw visual features.
The core intuition is a dynamic ``division of labor'': spatial anchors preserve crucial spatial relations and low-frequency features, while query tokens focus on complementary source-grid details required for video understanding, resolution-sensitive reasoning, and dense recognition.

\subsection{Spectral Bottlenecks and Spatial Aliasing in Token Compression}
\label{sec:spectral-analysis}
To understand how elastic visual-token compression changes the information carried by visual features, we analyze the compressed representations in the spatial-frequency domain.
This provides a natural lens for separating coarse layout from fine-grained detail: lower spatial frequencies capture slowly varying global structure, whereas higher spatial frequencies correspond to localized changes and detail-sensitive visual features.
We therefore use radial power spectral diagnostics to test how different frequency bands are suppressed, preserved, or emphasized by both our baselines and \ours.
The detailed mathematical protocol for the analysis provided below is in Appendix~\ref{app-sec:appendix_spectral_math}.

We first analyze the bottleneck induced by average pooling by evaluating the cumulative spectral concentration of the post-compression spatial grid.
This diagnostic avoids the scale ambiguity of raw input-output spectral transfer ratios, testing whether the compressed grid concentrates its spectral mass in the low-frequency baseband.
As shown in Figure~\ref{fig:spectral_a}, \ours accumulates spectral power more rapidly at low spatial frequencies than \(\text{M}^3\).
This indicates stronger low-frequency concentration within the spatial pool tokens of \ours.
Conversely, the broader accumulation in \(M^3\) reflects less selective low-pass behavior under spatial compression.
Because spatial decimation lowers the representable Nyquist range, this broad post-compression spectrum is consistent with spectral leakage or aliasing under aggressive downsampling \cite{oppenheim1999discrete,gonzalez1992digital,zhang2019making,azulay2019why}.

Query-only compression suffers from a complementary weakness.
MQT replaces explicit, grid-aligned spatial tokens with non-local learned summaries, and a nested dropout strategy enforces elasticity over this query sequence \cite{NestedDropout}.
While this elastic representation is highly flexible, it forces the queries to encode both the low-frequency layout and fine-grained semantic details without an underlying spatial anchor.
As demonstrated in Figure~\ref{fig:spectral_b}, MQT does not exhibit the same clear separation between low-frequency anchoring and higher-frequency features.
This structural weakness is empirically reflected in dense spatial grounding tasks: across the RefCOCO suite (Table~\ref{tab:detailed_video_refcoco_res_condensed}), \ours consistently outperforms MQT across all token budgets, achieving up to a \(+6.1\) point retention advantage at 64 tokens.

To mitigate these complementary bottlenecks, \ours adopts a dynamic division-of-labor strategy.
The spatial pool tokens provide an explicit low-frequency spatial anchor, while the pool-conditioned query pathway is encouraged to emphasize complementary visual information.
This design reduces the burden on query tokens to model the entire visual spectrum alone, while preserving an explicit spatial representation for layout-sensitive visual reasoning.

\subsection{Pool-Conditioned Query Resampling}
\label{sec:pool-self-attention-block}
\begin{figure*}[t]
    \centering
    \includegraphics[width=\textwidth]{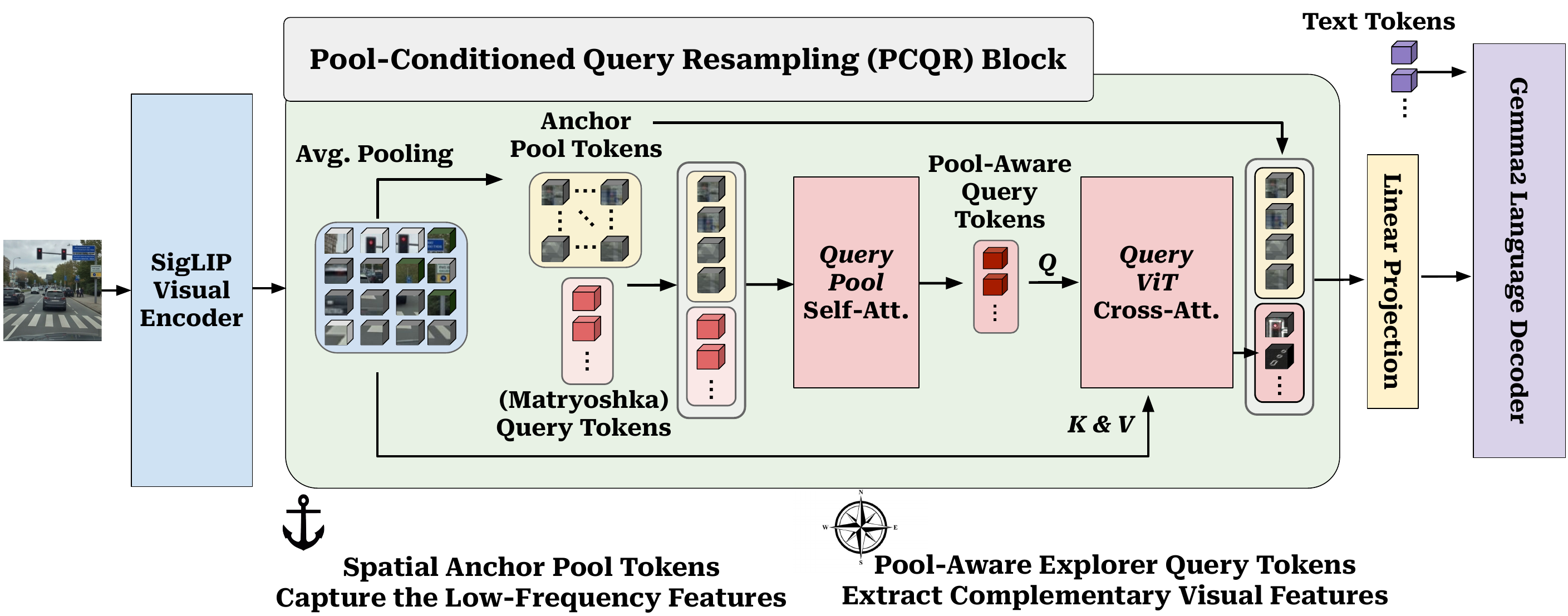}
\caption{\textbf{High-Level Overview of the \ours Architecture.} \ours dynamically divides the labor of visual feature extraction into a unified pipeline.
Uncompressed visual encoder features are first spatially pooled to create deterministic \textbf{2D Anchors} that secure the low-frequency geometric layout.
A supporting set of query tokens then undergoes \textbf{Pool-Conditioned Query Resampling (PCQR)}.
After interacting with the spatial anchors through PCQR, these queries act as \textbf{Semantic Explorers} that extract complementary information from the raw visual features.
The final concatenated representation provides an effective budget-aware context to the language decoder.}
    \label{fig:main_teaser}
\end{figure*}

To realize this spectral disentanglement, \ours couples low-frequency spatial anchoring with a complementary query pathway to explore a richer set of visual features.
As outlined in Figure~\ref{fig:main_teaser}, this is achieved through an efficient, sequential attention mechanism we term \textbf{Pool-Conditioned Query Resampling (PCQR)}.

Formally, let \(X_{v} \in \mathbb{R}^{N_{v} \times D}\) denote the uncompressed visual features extracted by the visual encoder.
Depending on the selected computational budget, PCQR applies budget-aware average pooling (\textit{e.g.}, \(2 \times 2\) or \(4 \times 4\)) to extract a grid-aligned spatial anchor representation.
We define these \textbf{``2D Anchor'' Pool Tokens} as \(P \in \mathbb{R}^{N_{p} \times D}\).

In parallel, a base set of unconditioned, learnable query tokens \(Q_{IN}\) undergoes nested dropout to support elastic query budgets.
With nested dropout, we sample a budget \(B\), keep only the first \(N_q=B-N_p\) query tokens after allocating anchors, and drop the rest.
Since earlier queries are active across more budgets, they learn a nested prefix structure that enables query truncation at inference without retraining.
To fuse the structural layout with the query pathway, we concatenate these sequences along the token dimension and process the joint representation through a unifying Query \(\leftrightarrow\) Pool Self-Attention block.
From the output of this block, we isolate the updated query sequence to obtain the \textbf{Pool-Aware Query Tokens}, denoted as \(Q_{PA}\).
This step explicitly conditions the query tokens on the pooled spatial anchors \textit{prior} to dense visual feature extraction, encouraging the query pathway to focus on complementary details absent from the spatial anchors.

After this conditioning step, the queries have access to the coarse spatial layout encoded by the anchors.
We then let them cross-attend to the full-resolution ViT features so that they can retrieve complementary visual information not represented in the pooled anchor grid.
Here, \(Q_{PA}\) serves as the queries (\(Q\)), while the raw, uncompressed features \(X_{v}\) act as the keys and values (\(K, V\)):
\begin{equation}
\label{eq:cross_attention}
Q_{SE} = \text{CrossAttn.}(Q=Q_{PA}, K=X_{v}, V=X_{v}).
\end{equation}
The resulting outputs are \textbf{``Semantic Explorer'' Query Tokens} (\(Q_{SE}\)), which capture complementary visual features.
Finally, the structural 2D anchors (\(P\)), the targeted semantic explorers (\(Q_{SE}\)), and the text tokens are concatenated and fed to the language decoder.

\subsection{Budget-Aware Piecewise Routing and Nested Dropout}
\label{sec:budget-aware-routing-method}

To effectively realize \ours across variable inference constraints, we further introduce a budget-aware piecewise routing strategy.
Let \(B\) denote the total allocated visual token budget for a given image or video fragment.
To balance spatial anchoring and semantic exploration, the routing mechanism dynamically determines the resolution of the spatial anchor \(P\) and the number of complementary query tokens \(N_q\) based on \(B\).

Specifically, we define two distinct routing regimes:
\begin{itemize}[leftmargin=0.5cm, itemsep=0.05cm, topsep=0.1cm]
    \item \textbf{Low Budgets (\(16 \le B < 64\)):} The uncompressed visual features are pooled into a \(4 \times 4\) spatial grid, yielding an anchor sequence of \(N_p = 16\) tokens. The remainder of the budget is filled by allocating \(N_q = B - 16\) query tokens.
    \item \textbf{Medium-to-High Budgets (\(64 \le B \le 256\)):} The model scales the spatial anchor to an \(8 \times 8\) grid, yielding \(N_p = 64\) structural tokens. The complementary query allocation becomes \(N_q = B - 64\).
\end{itemize}

These two anchor sizes match the evaluated budget range: \(4\times4\) preserves a minimal layout under extreme compression, while \(8\times8\) provides a richer spatial base at higher budgets without exhausting the token budget.
This allocation preserves an explicit spatial anchor at every budget while assigning the remaining tokens to the complementary query pathway.
At anchor-size budgets, this routing naturally reduces to a spatial-anchor representation; as the budget grows, additional query tokens provide source-grid detail.
%
%\vspace{-1.5ex}
\section{Results and Discussions}
\label{sec:experiments}
We now discuss the experimental evaluation of \ours and the baselines ($\text{M}^3$ and MQT) on the PaliGemma-2 evaluation suite spanning video understanding, dense recognition and vision-centric multimodal understanding tasks.
We evaluate \ours along three axes.
First, we measure aggregate performance retention across the benchmark suite to test whether \ours improves the global performance--token trade-off.
Second, we isolate benchmarks that aim for stressing the two bottlenecks identified in Section~\ref{sec:spectral-analysis}: spatial grounding, resolution-sensitive reasoning, and video understanding tasks.
Finally, we ablate the routing and attention design to verify that the gains arise from the proposed division of labor rather than from additional connector capacity alone.
\begin{table*}[t]
\centering
\caption{\textbf{Overall Mean Performance.}
Image represents an aggregation of the RefCOCO, Resolution-Sensitive, and General benchmarks. Video represents an aggregation of the video benchmarks included in the main evaluation. We report absolute macro-average raw scores alongside mean retention rates (\%) evaluated across 3 random seeds. Aggregated retention rates are calculated by taking the mean of individual benchmark retention rates. Best results per budget are \textbf{bolded}, and second-best are \underline{underlined}.}
\label{tab:overall_summary}
\begin{tabular}{@{}llccc@{}}
\toprule
\textbf{Modality} & \textbf{Visual Budget} & \textbf{M$^3$} & \textbf{MQT} & \textbf{\ours~(Ours)} \\
\midrule
\multirow{3}{*}{\textbf{Image}}
& 256 Tokens  & 67.6 (91.1\%) & \underline{69.1 (93.3\%)} & \textbf{70.4 (95.1\%)} \\
& 64 Tokens   & 66.5 (89.2\%) & \underline{68.1 (91.6\%)} & \textbf{70.1 (94.7\%)} \\
& 16 Tokens   & 63.9 (85.2\%) & \underline{64.3 (85.8\%)} & \textbf{64.8 (86.8\%)} \\
\midrule
\multirow{3}{*}{\textbf{Video}}
& 256 Tokens  & 50.6 (92.9\%) & \underline{51.4 (94.4\%)} & \textbf{53.1 (98.0\%)} \\
& 64 Tokens   & 50.3 (92.5\%) & \underline{50.9 (93.5\%)} & \textbf{53.1 (97.9\%)} \\
& 16 Tokens   & 49.8 (91.6\%) & \underline{51.2 (94.0\%)} & \textbf{51.6 (95.0\%)} \\
\bottomrule
\end{tabular}%}
\end{table*}

\begin{table*}[t]
\centering
\begingroup
\newcommand{\tblscore}[2]{\ensuremath{#1\,{\scriptstyle\textcolor{gray}{\pm #2}}}}
\newcommand{\tblbscore}[2]{\ensuremath{\mathbf{#1}\,{\scriptstyle\textcolor{gray}{\pm #2}}}}
\newcommand{\tbluscore}[2]{\underline{\ensuremath{#1\,{\scriptstyle\textcolor{gray}{\pm #2}}}}}
\newcommand{\methodhead}[1]{\makecell[c]{\textbf{#1}\\[-3pt]\scriptsize\vphantom{\textit{(Ours)}}}}
\newcommand{\parcelhead}{\makecell[c]{\textbf{PARCEL}\\[-3pt]\footnotesize\textbf{\textit{(Ours)}}}}
\newcommand{\benchhead}{\makecell[l]{\textbf{Benchmark}\\[-3pt]\scriptsize\vphantom{\textit{(Ours)}}}}
\caption{\textbf{Detailed Performance on Video, Image Segmentation, and Resolution-Sensitive Benchmarks.}
Raw scores are means over three seeds, with standard deviation shown in gray. Video and Resolution-Sensitive blocks show the top-3 compression-sensitive splits, selected by the largest 16-token retention drop and RefCOCO rows aggregate over splits. Mean Retention rows summarize each block, with RefCOCO computed over all RefCOCO splits. Vanilla PG2 is shaded as the uncompressed reference. Best are \textbf{bolded}, second-best are \underline{underlined}.}
\label{tab:detailed_video_refcoco_res_condensed}
\setlength{\tabcolsep}{2.5pt}
\renewcommand{\arraystretch}{1.08}
\resizebox{\textwidth}{!}{%
\begin{tabular}{@{}l >{\columncolor[gray]{0.92}}c ccc ccc ccc@{}}
\toprule
& \multicolumn{1}{c}{} 
& \multicolumn{3}{c}{\textbf{256 Visual Tokens}} 
& \multicolumn{3}{c}{\textbf{64 Visual Tokens}} 
& \multicolumn{3}{c}{\textbf{16 Visual Tokens}} \\
\cmidrule(lr){3-5} \cmidrule(lr){6-8} \cmidrule(lr){9-11}
\benchhead & \methodhead{PG2}
& \methodhead{M$^3$} & \methodhead{MQT} & \parcelhead
& \methodhead{M$^3$} & \methodhead{MQT} & \parcelhead
& \methodhead{M$^3$} & \methodhead{MQT} & \parcelhead \\
\midrule

\multicolumn{11}{@{}l}{\textit{Video Understanding Benchmarks (Top 3 Most Compression-Sensitive)}} \\
\midrule
ActivityNet-CAP & 43.7 
& \tblscore{36.1}{1.6} 
& \tbluscore{37.2}{1.0} 
& \tblbscore{41.5}{0.9} 
& \tblscore{36.5}{0.2} 
& \tbluscore{37.1}{1.3} 
& \tblbscore{40.5}{1.5} 
& \tblscore{36.5}{1.0} 
& \tbluscore{37.4}{0.9} 
& \tblbscore{38.9}{0.4} \\

ActivityNet-QA & 53.3 
& \tbluscore{51.5}{0.3} 
& \tblscore{50.0}{0.8} 
& \tblbscore{52.4}{0.7} 
& \tbluscore{51.2}{0.1} 
& \tblscore{49.8}{0.8} 
& \tblbscore{52.7}{0.5} 
& \tbluscore{50.4}{0.3} 
& \tbluscore{50.4}{0.9} 
& \tblbscore{50.6}{0.1} \\

MSRVTT-Cap & 70.6 
& \tblscore{64.2}{0.9} 
& \tbluscore{66.8}{2.0} 
& \tblbscore{68.6}{1.6} 
& \tblscore{63.4}{2.7} 
& \tbluscore{64.5}{2.2} 
& \tblbscore{68.9}{1.8} 
& \tblscore{62.4}{1.1} 
& \tbluscore{65.7}{1.8} 
& \tblbscore{66.7}{0.8} \\
\midrule
\textbf{Mean Retention} & -- 
& 90.1\% & \underline{91.2\%} & \textbf{96.8\%} 
& 89.7\% & \underline{89.9\%} & \textbf{96.3\%} 
& 88.8\% & \underline{91.1\%} & \textbf{92.8\%} \\
\midrule
\multicolumn{11}{@{}l}{\textit{Image Segmentation (RefCOCO)}} \\
\midrule
RefCOCO (Avg.) & 68.0 
& \tblscore{57.8}{0.3} 
& \tbluscore{61.0}{0.3} 
& \tblbscore{63.4}{0.2} 
& \tblscore{56.5}{0.3} 
& \tbluscore{60.3}{0.3} 
& \tblbscore{63.6}{0.3} 
& \tblscore{53.0}{0.2} 
& \tbluscore{56.2}{0.4} 
& \tblbscore{56.7}{0.1} \\

RefCOCO+ (Avg.) & 65.4 
& \tblscore{52.6}{0.5} 
& \tbluscore{55.5}{0.3} 
& \tblbscore{58.4}{0.2} 
& \tblscore{51.0}{0.4} 
& \tbluscore{54.2}{0.2} 
& \tblbscore{58.4}{0.2} 
& \tblscore{47.3}{0.3} 
& \tbluscore{50.0}{0.3} 
& \tblbscore{51.3}{0.3} \\

RefCOCO-g (Avg.) & 65.2 
& \tblscore{51.2}{0.4} 
& \tbluscore{54.6}{0.2} 
& \tblbscore{57.7}{0.1} 
& \tblscore{49.9}{0.5} 
& \tbluscore{53.5}{0.3} 
& \tblbscore{57.8}{0.2} 
& \tblscore{46.6}{0.2} 
& \tbluscore{50.2}{0.1} 
& \tblbscore{51.5}{0.1} \\
\midrule
\textbf{Mean Retention} & -- 
& 81.7\% & \underline{86.4\%} & \textbf{90.6\%} 
& 79.6\% & \underline{84.9\%} & \textbf{90.8\%} 
& 74.2\% & \underline{79.0\%} & \textbf{80.5\%} \\
\midrule
\multicolumn{11}{@{}l}{\textit{Resolution-Sensitive Tasks (Top 3 Most Compression-Sensitive)}} \\
\midrule
DocVQA (val) & 36.6 
& \tblscore{30.5}{0.3} 
& \tblbscore{33.4}{0.1} 
& \tbluscore{32.8}{0.3} 
& \tblscore{28.7}{0.1} 
& \tbluscore{31.2}{0.5} 
& \tblbscore{32.1}{0.4} 
& \tbluscore{25.9}{0.2} 
& \tblscore{24.4}{0.1} 
& \tblbscore{26.1}{0.1} \\

ChartQA (human) & 40.7 
& \tblscore{33.8}{0.3} 
& \tbluscore{35.6}{0.8} 
& \tblbscore{37.2}{0.6} 
& \tblscore{32.3}{1.0} 
& \tbluscore{34.1}{0.9} 
& \tblbscore{37.0}{0.3} 
& \tblscore{29.7}{0.7} 
& \tbluscore{30.7}{0.3} 
& \tblbscore{32.0}{1.0} \\

ChartQA (aug) & 72.5 
& \tblscore{64.1}{0.6} 
& \tbluscore{66.0}{0.5} 
& \tblbscore{66.2}{0.8} 
& \tblscore{63.0}{1.3} 
& \tbluscore{64.8}{0.2} 
& \tblbscore{66.3}{0.2} 
& \tblbscore{59.9}{0.8} 
& \tblscore{58.2}{0.9} 
& \tbluscore{58.8}{0.9} \\
\midrule
\textbf{Mean Retention} & -- 
& 84.9\% & \underline{90.0\%} & \textbf{90.8\%} 
& 81.6\% & \underline{86.2\%} & \textbf{90.0\%} 
& \underline{75.5\%} & 74.2\% & \textbf{77.1\%} \\
\bottomrule
\end{tabular}
}
\endgroup
\end{table*}
\subsection{Experimental Setup and Baselines}
We evaluate \ours along three axes: aggregate retention across 27 benchmarks, targeted analysis on compression-sensitive task groups, and ablations isolating the role of budget routing and Pool-Conditioned Query Resampling.
This lets us test both the global accuracy--efficiency trade-off and the two bottlenecks identified in Section~\ref{sec:spectral-analysis}: spatial grounding under query-only compression and fine-detail loss under rigid spatial pooling.
We implement \ours and all baselines using the PaliGemma-2~(PG2) 3B \cite{PG2}, consisting of a 2B Gemma-2 language decoder \cite{Gemma2} and a SigLIP-SO-400M vision encoder \cite{SigLIP}.
We choose PaliGemma-2 because it provides a compact yet capable open LVLM backbone with established support across the diverse task families needed for our study, including generic VQA, dense localization, resolution-sensitive document/chart reasoning, and video understanding.
The uncompressed model is denoted as Vanilla PG2 and serves as the reference for retention calculations.
We compare \ours against two elastic matryoshka compression paradigms: rigid spatial downsampling (\(\text{M}^3\) \cite{M3}) and non-local query resampling (MQT \cite{MQT}). For \(\text{M}^3\), we use the one-forward-pass-per-batch variant \cite{M3} to match training compute budgets.
For efficiency metrics, we follow prior works \cite{FastV, PyramidDrop}, with PaliGemma-2-specific adjustments detailed in Appendix~\ref{app-sec:flop-calculations}.
We emphasize that \ours introduces negligible training and inference overhead compared to MQT and M$^3$ since the PCQR block is very lightweight, especially compared to the ViT/LLM components.

\textbf{Benchmarks.} We evaluate on 27 vision-centric benchmarks spanning video understanding, dense spatial grounding, resolution-sensitive reasoning, and general multimodal comprehension.
Video tasks include ActivityNet-Cap~\cite{ActivityNetCap}, ActivityNet-QA~\cite{ActivityNetQA}, MSRVTT-Cap~\cite{MSRVTT-Cap}, MSRVTT-QA~\cite{MSVD1}, and MSVD-QA~\cite{MSVD1, MSVD2}.
Dense spatial grounding is evaluated on RefCOCO, RefCOCO+, and RefCOCO-g~\cite{RefCOCO1,RefCOCO2,RefCOCOG}.
Following the PaliGemma evaluation protocol \cite{PG}, we evaluate fine-grained visual reasoning using nine resolution-sensitive splits, including ChartQA human/augmented splits~\cite{ChartQA}, DocVQA~\cite{DocVQA}, InfoVQA~\cite{InfoVQA}, SciCap~\cite{hsu2021scicap}, ST-VQA~\cite{ST-VQA}, TextCaps~\cite{TextCaps}, TextVQA~\cite{TextVQA}, and WidgetCap~\cite{WidgetCap}.
To increase benchmark diversity, we additionally include GQA/xGQA~\cite{GQA,xGQA}, NLVR2/MARVL5~\cite{NLVR2,MARVL5}, and OCR-VQA~\cite{OCR-VQA}.
We also report the results on the remaining PaliGemma benchmarks in Appendix~\ref{app-sec:detailed-results} and the high-resolution results on a subset of these benchmarks in Appendix~\ref{app-sec:detailed-results-high-resolution}.
We adhere to the training settings of PaliGemma-2 \cite{PG2} whenever applicable.
Comprehensive dataset descriptions, evaluation metrics, and training hyperparameters are provided in Appendix~\ref{app-sec:benchmark-details} and Appendix~\ref{app-sec:implementation-details}.

\subsection{Evaluation Protocol and Design Choices}
We report absolute raw scores and retention relative to Vanilla PG2.
For method \(m\) at budget \(b\), retention is computed as \(100\times s_{m,b}/s_{\mathrm{PG2}}\), where \(s_{m,b}\) is the benchmark score and \(s_{\mathrm{PG2}}\) is the corresponding Vanilla PG2 score.
All table values are averaged over three random seeds, and aggregate retention is computed as the mean of per-benchmark retention rates.
For Table~\ref{tab:detailed_video_refcoco_res_condensed}, we report the top-3 compression-sensitive splits within the Video and Resolution-Sensitive groups, selected by the largest retention drop at the 16-token budget.
This focuses the analysis on settings where visual-token compression is most stressful.
For the remaining benchmarks, we refer to Appendix~\ref{app-sec:detailed-results}.

\subsection{Global Efficiency and Pareto Trade-offs}
Figure~\ref{fig:retention_flops_memory} summarizes the aggregate accuracy--efficiency trade-off across all 27 benchmarks.
Across all visual-token budgets, \ours achieves the highest mean retention among compressed models, improving over both MQT and \(\text{M}^3\).
Table~\ref{tab:overall_summary} shows that this advantage holds across both image and video domains.
In image benchmarks, \ours preserves \(95.1\%\) and \(94.7\%\) retention at 256 and 64 tokens, respectively, and remains best at 16 tokens with \(86.8\%\).
In video benchmarks, \ours preserves \(98.0\%\) and \(97.9\%\) retention at 256 and 64 tokens, and \(95.0\%\) under the 16-token constraint.
Since visual-token count controls decoder prefill and KV-cache cost, these gains translate into a stronger accuracy--efficiency trade-off relative to Vanilla PG2.

\subsection{Detailed Benchmark Analysis}
Next, we examine task groups that directly probe the bottlenecks identified in Section~\ref{sec:spectral-analysis}.
RefCOCO evaluates whether explicit spatial anchors improve dense localization, DocVQA and ChartQA stress fine-grained visual evidence, and video tasks test whether compression preserves action-relevant temporal information.

\textbf{Video Understanding.}
The video block of Table~\ref{tab:detailed_video_refcoco_res_condensed} shows that the same pattern extends to multi-frame inputs.
Averaging the top-3 compression-sensitive video tasks, \ours retains \(92.8\%\) of Vanilla PG2 performance at 16 tokens, outperforming MQT (\(91.1\%\)) and \(\text{M}^3\) (\(88.8\%\)).
These results further highlight the role of separating spatial anchoring from complementary feature extraction while compressing temporal visual evidence.

\textbf{Image Segmentation.}
The RefCOCO block of Table~\ref{tab:detailed_video_refcoco_res_condensed} directly tests the spatial grounding weakness of query-only compression.
Across the full RefCOCO suite, MQT drops to \(79.0\%\) mean retention at 16 tokens, whereas \ours retains explicit spatial anchors and achieves \(80.5\%\).
At 256 tokens, the gap becomes larger: \ours reaches \(90.6\%\) mean retention, outperforming MQT by \(\mathbf{+4.2}\) points and \(\text{M}^3\) by \(\mathbf{+8.9}\) points.

\textbf{Resolution-Sensitive Benchmarks.}
The resolution-sensitive block of Table~\ref{tab:detailed_video_refcoco_res_condensed} stresses fine-grained visual evidence through DocVQA and ChartQA.
Averaging the top-3 resolution-sensitive tasks at 16 tokens, \(\text{M}^3\) reaches \(75.5\%\) mean retention and MQT drops to \(74.2\%\), while \ours achieves the best retention at \(77.1\%\).
At this boundary budget, \ours operates through its spatial-anchor representation, showing that the anchor branch itself provides a stronger compressed visual base.
At larger budgets, \ours reaches \(90.0\%\) retention at 64 tokens from the expanded spatial base and \(90.8\%\) at 256 tokens once the queries become active.
\subsection{Ablations on Design Choices}
\label{sec:ablations}

\begin{table*}[t]
\centering
\caption{\textbf{Ablation Studies on Architecture and Routing.} Overall Mean Retention Rate (\%) across the 27-benchmark main evaluation suite. The first row is the full \ours configuration, whereas subsequent blocks isolate budget routing, attention design, and baseline-capacity controls. Unsupported budgets are marked as \textcolor{gray}{N/A}. Best or tied-best displayed values per budget are \textbf{bolded}.}
\label{tab:ablations}
%\resizebox{0.65\linewidth}{!}{%
\begin{tabular}{@{}c l c c c@{}}
\toprule
\textbf{Ablation} & \textbf{Model Configuration} & \textbf{256} & \textbf{64} & \textbf{16} \\
\midrule
\textbf{Full} 
& \textbf{\ours \textit{(Ours)}} 
& \textbf{95.6\%} & \textbf{95.3\%} & \textbf{88.3\%} \\
\midrule

\multirow{4}{*}{\makecell[c]{\textit{Budget}\\[-1pt]\textit{Routing}}}
& Average Pooling ($4\times4$) 
& \textcolor{gray}{N/A} & \textcolor{gray}{N/A} & 83.1\% \\
& Average Pooling ($2\times2$) 
& \textcolor{gray}{N/A} & 92.8\% & \textcolor{gray}{N/A} \\
& \ours w/ Fixed $4\times4$ Routing 
& 90.2\% & 89.6\% & \textbf{88.3\%} \\
& \ours w/ Fixed $2\times2$ Routing 
& \textbf{95.6\%} & 95.2\% & \textcolor{gray}{N/A} \\
\midrule

\multirow{2}{*}{\makecell[c]{\textit{Attention}\\[-1pt]\textit{Design}}}
& ViT-Only Cross-Attention 
& 95.2\% & \textbf{95.3\%} & 87.9\% \\
& Dual Cross-Attention (ViT + Pool) 
& 95.4\% & 95.2\% & 88.0\% \\
\midrule

\multirow{2}{*}{\makecell[c]{\textit{Baseline}\\[-1pt]\textit{Fairness}}}
& MQT w/ Self-Attention 
& 93.3\% & 92.5\% & 87.8\% \\
& M$^3$ w/ Self-Attention 
& 92.2\% & 90.4\% & 86.8\% \\
\bottomrule
\end{tabular}%
%}
\end{table*}

In this section, we quantify the effects of the critical building blocks of our method: (i) the budget-aware piecewise routing strategy from Section~\ref{sec:budget-aware-routing-method}, (ii) the Pool-Conditioned Query Resampling (PCQR) mechanism from Section~\ref{sec:pool-self-attention-block}, and (iii) enhanced baseline configurations to isolate the impact of our ``division of labor'' strategy.
All models are evaluated by mean retention across 27 benchmarks.

\textbf{Impact of Budget-Aware Routing.}
We hypothesize that spatial anchors must scale relative to the overall token budget.
The Budget Routing block of Table~\ref{tab:ablations} validates this necessity.
Relying solely on spatial anchors without query tokens, as in the Average Pooling baselines, severely caps performance, achieving only \(83.1\%\) and \(92.8\%\) retention at 16 and 64 tokens, respectively. 
Conversely, fixing the spatial anchor size regardless of the total budget restricts the model performance. For instance, maintaining a highly compressed 16-token anchor (Fixed $4\times4$ Routing) across all budgets causes performance to stagnate at $90.2\%$, even given a generous 256-token allowance.
In this regime, the query tokens are forced to bear the majority of the representational burden, attempting to reconstruct mid-frequency structural details that a larger anchor would have naturally preserved.
Similarly, the Average Pooling ($2 \times 2$) baseline improves upon this retention but cannot accommodate inference budgets below 64 tokens, as the number of pooling tokens dictates the minimum token count.
By implementing dynamic routing where the model scales to a 64-token anchor at higher budgets while falling back to 16 tokens under severe constraints, \ours optimally balances spatial anchors and semantic exploration.
This achieves peak retention across all constraints ($95.6\%$ at 256 tokens, $95.3\%$ at 64 tokens, and $88.3\%$ at 16 tokens).

\textbf{Efficacy of Pool-Conditioned Query Resampling.} 
The Attention Design block of Table~\ref{tab:ablations} studies the architectural flow of information between the pool tokens, query tokens, and raw visual features.
We compare our sequential PCQR module against two alternatives: a ``ViT-Only'' mechanism where queries cross-attend to the raw visual features without attending to spatial anchors, and a ``Dual Cross-Attention'' mechanism where queries first cross-attend to the pool tokens and then the ViT features.
Our final design first performs a full self-attention between the query tokens and the spatial anchors.
This allows the tokens to become structurally aware of one another, enabling the model to better allocate feature sampling based on the anchor's coverage.
These pool-aware query tokens then cross-attend to the raw visual features to sample the missing, complementary details not captured by the spatial anchors.
Consequently, our sequential design (Pool Self-Attn $\rightarrow$ ViT Cross-Attn) achieves the best or tied-best retention across budgets.
At the 256-token budget, PCQR reaches \(95.6\%\) retention, compared to \(95.2\%\) for ViT-only cross-attention and \(95.4\%\) for dual cross-attention.
The performance delta supports our design choice: for the division of labor to effectively work, the queries must be \textit{pool-aware}.
By conditioning the queries on the spatial anchors prior to visual feature extraction, the model guides them away from redundant low-frequency features, reserving their capacity for complementary visual information, as also reflected in Figure~\ref{fig:spectral_analysis}.

\textbf{Isolating the Division of Labor (Baseline Fairness).} 
A natural concern is whether the observed gains arise from added learnable parameters, \textit{i.e.}, the extra attention blocks, rather than the structural design itself.
To ensure baseline fairness, we upgrade both MQT and M$^3$ by adding comparable self-attention blocks, matching the parameter count and depth of \ours.
As shown in the Baseline Fairness block of Table~\ref{tab:ablations}, merely scaling capacity does not resolve the foundational bottlenecks of the baselines.
Upgraded MQT reaches \(93.3\%\) at 256 tokens, but still falls short of \ours (\(95.6\%\)) because it inherently lacks spatial anchors.
Similarly, upgraded M$^3$ reaches \(92.2\%\) at 256 tokens and remains below \ours across all budgets, indicating that added self-attention does not overcome the bottlenecks caused by rigid spatial downsampling.
These results support that the gains of \ours are not a product of added complexity alone, but rather from \ours's dynamic division of labor.

\section{Conclusion}
\label{ref:conclusion}

Large Vision-Language Models face severe computational bottlenecks during inference, and existing elastic compression paradigms force a stark trade-off between spectral aliasing and degraded spatial grounding.
To resolve this, we introduced \ours, a novel architecture that dynamically partitions the labor of visual feature extraction.
By coupling spatial anchors with pool-conditioned semantic queries, \ours disentangles low-frequency geometric layouts from high-frequency visual details.
Extensive evaluations across 27 vision-centric benchmarks demonstrate that this spectral partitioning establishes a new performance-efficiency Pareto frontier.
Through budget-aware routing, \ours sustains robust dense recognition, temporal reasoning, and resolution-sensitive performance even under significant 16-token constraints.
\ours preserves the highly desirable ``train once, deploy anywhere'' paradigm without sacrificing performance, providing a highly efficient foundation for ubiquitous LVLM deployment.

\textbf{Acknowledgements.} The authors would like to thank
(in alphabetic order of first name) Diego Martin Arroyo, Luca Zanella, Theo Uscidda, Yannick Strümpler for helpful comments, feedback and support throughout the project.

\bibliography{main}
\appendix
\newpage
\tableofcontents
\section{Spectral Analysis Protocol}
\label{app-sec:appendix_spectral_math}

We use spectral diagnostics to analyze how different visual-token compression mechanisms allocate spatial-frequency power. Spatial pooling can be viewed as a downsampling operation: reducing a feature grid from \(H\times W\) to \(H'\times W'\) lowers the maximum representable spatial frequency, and components above the reduced Nyquist limit may fold or leak into lower frequencies if they are not attenuated before decimation \cite{ShannonNoise,oppenheim1999discrete,zhang2019making}. Our goal is not to reconstruct the original feature map, but to characterize the extent to which compressed spatial tokens capture low-frequency features and query tokens emphasize higher-frequency source-grid detail beyond the spatial anchor pool tokens.

\paragraph{Feature grids.} Let \(\mathbf{M}\in\mathbb{R}^{H_M\times W_M\times C}\) denote a visual feature grid, where \(H_M\times W_M\) is the native spatial resolution and \(C\) is the channel dimension. All spectra are computed on native feature grids before projection into the language-model embedding space. Pooled grids are not upsampled before Fourier analysis.

We first remove the spatially constant component of each channel:

\begin{equation}
    \tilde{\mathbf{M}}_{h,w,c}
    =
    \mathbf{M}_{h,w,c}
    -
    \frac{1}{H_MW_M}
    \sum_{h'=0}^{H_M-1}
    \sum_{w'=0}^{W_M-1}
    \mathbf{M}_{h',w',c}.
\end{equation}

%We then compute a coefficient-normalized 2D Discrete Fourier Transform:
%We then compute the 2D Discrete Fourier Transform over the spatial dimensions and explicitly divide the Fourier coefficients by the number of spatial locations, \(H_MW_M\):
We then compute the forward normalized 2D Discrete Fourier Transform.
For notational brevity, we first define the complex exponential basis term $\mathcal{E}_{h,w}(u,v)$ as:
\begin{equation}
    \mathcal{E}_{h,w}(u,v) = \exp\left[ -2\pi i \left( \frac{uh}{H_M} + \frac{vw}{W_M} \right) \right].
\end{equation}
The transform is then compactly given by:
\begin{equation}
    \widehat{\mathbf{M}}_c(u,v) = \frac{1}{H_MW_M} \sum_{h=0}^{H_M-1} \sum_{w=0}^{W_M-1} \tilde{\mathbf{M}}_{h,w,c} \mathcal{E}_{h,w}(u,v).
\end{equation}

This normalization prevents larger spatial grids from producing larger Fourier magnitudes solely because they contain more samples.
Under this convention, Parseval's relation gives:

\begin{equation}
    \sum_{u,v}
    \left|
    \widehat{\mathbf{M}}_c(u,v)
    \right|^2
    =
    \frac{1}{H_MW_M}
    \sum_{h,w}
    \left|
    \tilde{\mathbf{M}}_{h,w,c}
    \right|^2,
\end{equation}

so the summed spectral power corresponds to the average spatial AC energy per feature location. This makes spectra scale-consistent across grids of different spatial resolutions \cite{oppenheim1999discrete,bracewell1989fourier,gonzalez1992digital}.

Finally, we compute the channel-averaged power spectrum, which yields power spectral density (PSD):
\begin{equation}
    S_{\mathbf{M}}(u,v)
    =
    \frac{1}{C}
    \sum_{c=1}^{C}
    \left|
    \widehat{\mathbf{M}}_c(u,v)
    \right|^2 .
\end{equation}

\paragraph{Radial mean power.}
To obtain one-dimensional frequency profiles for our visualizations, we collapse the 2D Fourier spectrum into a radial profile.
After applying the standard FFT shift, the DC component is placed at the frequency origin, and each Fourier coefficient can be indexed by zero-centered spatial-frequency coordinates:

\begin{equation}
\begin{split}
    u &\in \left\{ -\left\lfloor\frac{H_M}{2}\right\rfloor, \ldots, \left\lceil\frac{H_M}{2}\right\rceil - 1 \right\}, \\
    v &\in \left\{ -\left\lfloor\frac{W_M}{2}\right\rfloor, \ldots, \left\lceil\frac{W_M}{2}\right\rceil - 1 \right\}.
\end{split}
\end{equation}

The radial frequency of a coefficient is its Euclidean distance from the origin in this frequency plane:

\begin{equation}
    \rho(u,v)=\sqrt{u^2+v^2}.
\end{equation}

Radial averaging is reliable only for frequency rings that are fully represented inside the finite 2D Fourier grid.
We therefore restrict comparisons to the largest centered circle that fits inside this grid, \textit{i.e.,} the inscribed Nyquist radius:
\begin{equation}
    r_{\max}(\mathbf{M})
    =
    \frac{1}{2}
    \min(H_M,W_M).
\end{equation}
This native-grid cutoff is the reason for different x-axis ranges in Figure~\ref{fig:spectral_analysis}.
To exemplify, a \(16\times16\) source grid has an inscribed radial Nyquist limit of \(r_{\max}=8\), whereas an \(8\times8\) pooled grid has \(r_{\max}=4\). 
Since pooled tokens are analyzed on their native compressed grid without upsampling, their radial profiles terminate at the pooled-grid limit.
Query-attention weighted ViT features (denoted by \ours--Query and MQT) are computed on the original source grid and therefore retain the higher source-grid frequency support up to the source-grid Nyquist limit.

To form discrete radial bins, we group Fourier coefficients into unit-width rings around the origin.
The radial bin at frequency radius \(r\) is given by:

\begin{equation}
    \mathcal{R}_r = \left\{ (u,v) : 
    \begin{aligned}
        &r-\frac{1}{2} \leq \rho(u,v) < r+\frac{1}{2}, \\
        &\rho(u,v)\leq r_{\max}(\mathbf{M})
    \end{aligned}
    \right\}.
\end{equation}

The radial mean power is then the average PSD value within each ring:

\begin{equation}
    P_{\mathbf{M}}(r)
    =
    \frac{1}{|\mathcal{R}_r|}
    \sum_{(u,v)\in\mathcal{R}_r}
    S_{\mathbf{M}}(u,v).
\end{equation}

This measures the average spectral power per Fourier coefficient at radius \(r\).
Unlike annular energy, it does not give extra weight to high-frequency rings merely because they contain more Fourier coefficients.

For dataset-level curves, we compute \(P_{\mathbf{M}}(r)\) per sample and then average the resulting radial profiles:

\begin{equation}
    \bar{P}_{\mathbf{M}}(r)
    =
    \mathbb{E}_{x\sim\mathcal{D}}
    \left[
    P_{\mathbf{M}(x)}(r)
    \right].
\end{equation}

\paragraph{Cumulative spectral concentration.}
Figure~\ref{fig:spectral_analysis}(a) analyzes the extent to which the compressed spatial tokens capture the low-frequency baseband.
For a compressed spatial grid
\(\mathbf{Y}\in\mathbb{R}^{H_{\mathrm{out}}\times W_{\mathrm{out}}\times C}\), we first normalize its dataset-averaged radial mean-power profile:

\begin{equation}
    \hat{P}_{\mathbf{Y}}(r)
    =
    \frac{
    \bar{P}_{\mathbf{Y}}(r)
    }{
    \sum_{k=1}^{r_{\max}(\mathbf{Y})}
    \bar{P}_{\mathbf{Y}}(k)
    +
    \epsilon
    },
\end{equation}

where \(\epsilon=10^{-15}\) is used for numerical stability.
We then define the cumulative spectral concentration as:

\begin{equation}
    C_{\mathbf{Y}}(r)
    =
    \sum_{k=1}^{r}
    \hat{P}_{\mathbf{Y}}(k).
\end{equation}

A steeply rising \(C_{\mathbf{Y}}(r)\) indicates that most normalized spectral power is concentrated in low spatial frequencies, implying a heavier focus on capturing low-frequency visual features.
On the other hand, a slower-rising curve indicates that spectral power is distributed more broadly over the available frequency range, implying that the compressed tokens capture a more spread-out band of feature frequencies.

\paragraph{Normalized radial mean power.}
Figure~\ref{fig:spectral_analysis}(b) visualizes the normalized radial mean-power directly:

\begin{equation}
    \hat{P}_{\mathbf{M}}(r)
    =
    \frac{
    \bar{P}_{\mathbf{M}}(r)
    }{
    \sum_{k=1}^{r_{\max}(\mathbf{M})}
    \bar{P}_{\mathbf{M}}(k)
    +
    \epsilon
    }.
\end{equation}

This profile measures relative per-mode frequency concentration independent of total feature magnitude.
When the target feature map is the pooled spatial grid, we set \(\mathbf{M}=\mathbf{Y}\) in the definition above and obtain \(\hat{P}_{\mathbf{Y}}(r)\).
For spatial pool tokens, \(\hat{P}_{\mathbf{Y}}(r)\) is evaluated only up to the pooled-grid Nyquist radius.
For query-based compressors such as MQT \cite{MQT} and \ours, query tokens do not possess a native 2D structure because they are learned non-spatial sequence tokens where nested dropout enforces elasticity over this query sequence \cite{NestedDropout}.
We therefore analyze their attention-weighted footprint on the ViT output features.

Let
\(\mathbf{A}\in\mathbb{R}^{N_q\times HW}\)
denote the post-softmax query-to-visual attention matrix from \(N_q\) query tokens to the ViT output tokens, averaged over attention heads.
For query \(q\), the vector \(\mathbf{A}_{q,:}\in\mathbb{R}^{HW}\) contains its attention weights over all spatial ViT tokens.
We reshape this vector into a 2D attention map \(\mathbf{A}_{q}\in\mathbb{R}^{H\times W}\) and rescale it to unit spatial mean:

\begin{equation}
    \tilde{\mathbf{A}}_{q,h,w}
    =
    \frac{
    \mathbf{A}_{q,h,w}
    }{
    \frac{1}{HW}
    \sum_{h'=0}^{H-1}
    \sum_{w'=0}^{W-1}
    \mathbf{A}_{q,h',w'}
    +
    \epsilon
    }.
\end{equation}

The query-attended source feature map is then:

\begin{equation}
    \mathbf{Z}^{(q)}_{h,w,c}
    =
    \mathbf{X}_{h,w,c}
    \tilde{\mathbf{A}}_{q,h,w},
\end{equation}

where \(\mathbf{X}\in\mathbb{R}^{H\times W\times C}\) is the source visual grid.
We compute the PSD of each \(\mathbf{Z}^{(q)}\) and average across queries:

\begin{equation}
    S_{\mathbf{Z}}(u,v)
    =
    \frac{1}{N_q}
    \sum_{q=1}^{N_q}
    S_{\mathbf{Z}^{(q)}}(u,v).
\end{equation}

We then apply the same radial mean-power computation defined above to \(S_{\mathbf{Z}}(u,v)\), followed by the same normalization, to obtain \(\hat{P}_{\mathbf{Z}}(r)\).
The resulting \(\hat{P}_{\mathbf{Z}}(r)\) is interpreted as the normalized radial mean-power profile of the ViT output feature regions weighted by query-token attention, rather than as a direct Fourier transform of the query-token sequence.

Together, these two diagnostics test the spectral role of each token family across both our baselines, M$^3$ and MQT, and our proposed method, \ours.
The cumulative curve in Figure~\ref{fig:spectral_analysis}(a) measures how quickly compressed spatial tokens concentrate their power into low frequencies. 
The normalized profiles in Figure~\ref{fig:spectral_analysis}(b) compare pool tokens and query-attended ViT output tokens, allowing us to assess whether spatial tokens serve as low-frequency anchors while query tokens retain access to higher-frequency features.
\section{Additional Experimental Results and Discussions}
\label{app-sec:additional-results}

This section provides the benchmark-level breakdown supporting the aggregate results in the main paper.
In Appendix~\ref{app-sec:detailed-results-high-resolution}, we further evaluate the compressed models after high-resolution \(448\times448\) pretraining, corresponding to the Stage-2 high-resolution setting of PaliGemma variants~\cite{PG,PG2}.
In Appendix~\ref{app-sec:detailed-results-high-resolution}, we further evaluate the compressed models after high-resolution \(448\times448\) pretraining (corresponds to the Stage 2 pretraining of PaliGemma variants \cite{PG, PG2}), providing an additional stress test for settings where visual detail is especially important.
Together, these results offer a more complete view of where \ours provides the largest gains and where different compression strategies behave similarly.

\subsection{Detailed Results Across All PaliGemma-2 Benchmarks}
\label{app-sec:detailed-results}
Tables~\ref{tab:detailed_video}--\ref{tab:detailed_general} provide the \(224\times224\) PaliGemma-2 evaluation results across video understanding, image segmentation, resolution-sensitive, and general vision-language benchmarks.
All scores are reported as the mean over three random seeds, with standard deviations shown in gray.
To keep the tables readable, per-benchmark retention values are omitted, while aggregate retention relative to Vanilla PG2 is reported in the final row of each table.

Overall, \ours provides the strongest aggregate retention on the task groups most affected by visual-token compression.
On video understanding benchmarks, \ours achieves the highest mean retention at all token budgets, reaching \(98.0\%\), \(97.9\%\), and \(95.0\%\) retention at 256, 64, and 16 tokens, respectively.
This indicates that the proposed hybrid connector preserves temporal visual evidence more effectively than both \(\text{M}^3\) and MQT under compression.

The advantage is even more pronounced on image segmentation benchmarks.
Across the RefCOCO suite, \ours consistently outperforms both baselines at every budget, achieving \(90.6\%\) retention at 256 tokens and \(90.8\%\) at 64 tokens, compared to \(86.4\%\) and \(84.9\%\) for MQT.
Even at the highly constrained 16-token setting, \ours remains the strongest model with \(80.5\%\) retention.
These results support the role of explicit 2D spatial anchors in preserving layout-sensitive information.

On resolution-sensitive benchmarks, \ours achieves the strongest aggregate retention at 256 and 64 tokens.
At 256 tokens, \ours reaches \(96.7\%\) mean retention, improving over MQT (\(96.3\%\)) and \(\text{M}^3\) (\(94.8\%\)).
At 64 tokens, the gap becomes larger, with \ours retaining \(95.7\%\), compared to \(94.1\%\) for MQT and \(92.3\%\) for \(\text{M}^3\).
At the extreme 16-token budget, \(\text{M}^3\) obtains the highest aggregate retention on this group, while \ours remains competitive and outperforms MQT.

For general vision-language benchmarks, all compressed models retain a large fraction of Vanilla PG2 performance, suggesting that many of these tasks are less sensitive to aggressive visual-token compression.
Even in this saturated regime, \ours achieves the best aggregate retention at 256 and 64 tokens, reaching \(99.4\%\) and \(99.2\%\), respectively.
At 16 tokens, \(\text{M}^3\) obtains the highest aggregate retention, while \ours remains competitive.
Together, these detailed results show that the gains of \ours are concentrated where compression is most challenging, video understanding, spatial grounding, and resolution-sensitive reasoning, while remaining competitive on broader multimodal benchmarks.

\paragraph{VATEX validation results.}
Finally, we also evaluate VATEX~\cite{VATEX}, but exclude it from aggregate summaries because we observe very high variance on the validation split across all methods and the official test set is not publicly available.
For completeness, at 256 tokens, \(\text{M}^3\), MQT, and \ours obtain \(78.7_{\pm 1.1}\), \(77.9_{\pm 3.6}\), and \(78.4_{\pm 1.8}\), respectively, compared to the Vanilla PG2 reference of \(80.5\).
At 64 tokens, the corresponding scores are \(78.8_{\pm 1.5}\), \(79.6_{\pm 1.9}\), and \(79.5_{\pm 2.6}\), while at 16 tokens they are \(79.6_{\pm 1.9}\), \(77.2_{\pm 1.1}\), and \(77.8_{\pm 2.7}\).
Because the observed method differences are comparable to the seed-level variation on this validation-only benchmark, we report these numbers for transparency but exclude VATEX from aggregate retention calculations.

\newcommand{\appscore}[2]{\ensuremath{#1\,{\scriptstyle\textcolor{gray}{\pm #2}}}}
\newcommand{\appbscore}[2]{\ensuremath{\mathbf{#1}\,{\scriptstyle\textcolor{gray}{\pm #2}}}}
\newcommand{\appuscore}[2]{\underline{\ensuremath{#1\,{\scriptstyle\textcolor{gray}{\pm #2}}}}}
\newcommand{\appmethodhead}[1]{\makecell[c]{\textbf{#1}\\[-2pt]\scriptsize\vphantom{\textbf{\textit{(Ours)}}}}}
\newcommand{\apparcelhead}{\makecell[c]{\textbf{PARCEL}\\[-2pt]\footnotesize\textbf{\textit{(Ours)}}}}
\newcommand{\appbenchhead}{\makecell[l]{\textbf{Benchmark}\\[-2pt]\scriptsize\vphantom{\textbf{\textit{(Ours)}}}}}
\begin{table*}[t]
\centering
\begingroup
\providecommand{\appscore}[2]{\ensuremath{#1\,{\scriptstyle\textcolor{gray}{\pm #2}}}}
\providecommand{\appbscore}[2]{\ensuremath{\mathbf{#1}\,{\scriptstyle\textcolor{gray}{\pm #2}}}}
\providecommand{\appuscore}[2]{\underline{\ensuremath{#1\,{\scriptstyle\textcolor{gray}{\pm #2}}}}}
\providecommand{\appmethodhead}[1]{\makecell[c]{\textbf{#1}\\[-2pt]\scriptsize\vphantom{\textbf{\textit{(Ours)}}}}}
\providecommand{\apparcelhead}{\makecell[c]{\textbf{PARCEL}\\[-2pt]\footnotesize\textbf{\textit{(Ours)}}}}
\providecommand{\appbenchhead}{\makecell[l]{\textbf{Benchmark}\\[-2pt]\scriptsize\vphantom{\textbf{\textit{(Ours)}}}}}
\caption{\textbf{Detailed Performance on Video Understanding Benchmarks.}
Raw scores are reported as the mean over three random seeds, with standard deviation shown in gray. Per-benchmark retention values are omitted for readability and aggregate retention relative to Vanilla PG2 is reported in the final row. Vanilla PG2 is shaded in grey and serves as the uncompressed reference. Best results per budget are \textbf{bolded}, and second-best are \underline{underlined}.}
\label{tab:detailed_video}
{
\setlength{\tabcolsep}{3.5pt}
\renewcommand{\arraystretch}{1.18}
\resizebox{\textwidth}{!}{%
\begin{tabular}{@{}l >{\columncolor[gray]{0.92}}c ccc ccc ccc@{}}
\toprule
& \multicolumn{1}{c}{} 
& \multicolumn{3}{c}{\textbf{256 Visual Tokens}} 
& \multicolumn{3}{c}{\textbf{64 Visual Tokens}} 
& \multicolumn{3}{c}{\textbf{16 Visual Tokens}} \\
\cmidrule(lr){3-5} \cmidrule(lr){6-8} \cmidrule(lr){9-11}
\appbenchhead & \appmethodhead{PG2}
& \appmethodhead{M$^3$} & \appmethodhead{MQT} & \apparcelhead
& \appmethodhead{M$^3$} & \appmethodhead{MQT} & \apparcelhead
& \appmethodhead{M$^3$} & \appmethodhead{MQT} & \apparcelhead \\
\midrule

ActivityNet-CAP 
& 43.7 
& \appscore{36.1}{1.6} 
& \appuscore{37.2}{1.0} 
& \appbscore{41.5}{0.9} 
& \appscore{36.5}{0.2} 
& \appuscore{37.1}{1.3} 
& \appbscore{40.5}{1.5} 
& \appscore{36.5}{1.0} 
& \appuscore{37.4}{0.9} 
& \appbscore{38.9}{0.4} \\

ActivityNet-QA 
& 53.3 
& \appuscore{51.5}{0.3} 
& \appscore{50.0}{0.8} 
& \appbscore{52.4}{0.7} 
& \appuscore{51.2}{0.1} 
& \appscore{49.8}{0.8} 
& \appbscore{52.7}{0.5} 
& \appuscore{50.4}{0.3} 
& \appuscore{50.4}{0.9} 
& \appbscore{50.6}{0.1} \\

MSRVTT-Cap 
& 70.6 
& \appscore{64.2}{0.9} 
& \appuscore{66.8}{2.0} 
& \appbscore{68.6}{1.6} 
& \appscore{63.4}{2.7} 
& \appuscore{64.5}{2.2} 
& \appbscore{68.9}{1.8} 
& \appscore{62.4}{1.1} 
& \appuscore{65.7}{1.8} 
& \appbscore{66.7}{0.8} \\

MSRVTT-QA 
& 41.5 
& \appscore{40.2}{0.1} 
& \appuscore{41.5}{0.6} 
& \appbscore{42.8}{0.6} 
& \appscore{39.9}{0.2} 
& \appuscore{40.9}{0.4} 
& \appbscore{43.0}{0.6} 
& \appscore{39.5}{0.1} 
& \appuscore{40.1}{0.6} 
& \appbscore{41.9}{0.1} \\

MSVD-QA 
& 62.7 
& \appuscore{60.8}{1.2} 
& \appbscore{61.4}{1.0} 
& \appscore{60.5}{1.8} 
& \appuscore{60.7}{1.2} 
& \appbscore{62.2}{0.8} 
& \appscore{60.6}{1.6} 
& \appuscore{60.4}{0.3} 
& \appbscore{62.6}{0.7} 
& \appscore{59.8}{0.3} \\

\midrule
\textbf{Mean Retention} 
& -- 
& 92.9\% & \underline{94.4\%} & \textbf{98.0\%} 
& 92.5\% & \underline{93.5\%} & \textbf{97.9\%} 
& 91.6\% & \underline{94.0\%} & \textbf{95.0\%} \\
\bottomrule
\end{tabular}
}
}
\endgroup
\end{table*}

\begin{table*}[t]
\centering
\caption{\textbf{Detailed Performance on Image Segmentation (RefCOCO) Benchmarks.}
Raw scores are reported as the mean over three random seeds, with standard deviation shown in gray. Per-benchmark retention values are omitted for readability and aggregate retention relative to Vanilla PG2 is reported in the final row. Vanilla PG2 is shaded in grey and serves as the uncompressed reference. Best results per budget are \textbf{bolded}, and second-best are \underline{underlined}.}
\label{tab:detailed_refcoco}
{
\setlength{\tabcolsep}{3.5pt}
\renewcommand{\arraystretch}{1.18}
\resizebox{\textwidth}{!}{%
\begin{tabular}{@{}l >{\columncolor[gray]{0.92}}c ccc ccc ccc@{}}
\toprule
& \multicolumn{1}{c}{} 
& \multicolumn{3}{c}{\textbf{256 Visual Tokens}} 
& \multicolumn{3}{c}{\textbf{64 Visual Tokens}} 
& \multicolumn{3}{c}{\textbf{16 Visual Tokens}} \\
\cmidrule(lr){3-5} \cmidrule(lr){6-8} \cmidrule(lr){9-11}
\appbenchhead & \appmethodhead{PG2}
& \appmethodhead{M$^3$} & \appmethodhead{MQT} & \apparcelhead
& \appmethodhead{M$^3$} & \appmethodhead{MQT} & \apparcelhead
& \appmethodhead{M$^3$} & \appmethodhead{MQT} & \apparcelhead \\
\midrule

RefCOCO (testA) 
& 71.8 
& \appscore{59.6}{0.4} 
& \appuscore{63.4}{0.2} 
& \appbscore{65.2}{0.2} 
& \appscore{58.2}{0.3} 
& \appuscore{62.6}{0.2} 
& \appbscore{65.4}{0.2} 
& \appscore{54.5}{0.2} 
& \appbscore{57.7}{0.4} 
& \appuscore{57.6}{0.1} \\

RefCOCO (testB) 
& 65.2 
& \appscore{56.1}{0.2} 
& \appuscore{58.6}{0.4} 
& \appbscore{61.7}{0.4} 
& \appscore{55.0}{0.4} 
& \appuscore{58.1}{0.5} 
& \appbscore{61.9}{0.4} 
& \appscore{51.9}{0.2} 
& \appuscore{55.0}{0.5} 
& \appbscore{56.1}{0.1} \\

RefCOCO (val) 
& 67.2 
& \appscore{57.7}{0.4} 
& \appuscore{61.0}{0.2} 
& \appbscore{63.3}{0.1} 
& \appscore{56.4}{0.3} 
& \appuscore{60.2}{0.2} 
& \appbscore{63.5}{0.3} 
& \appscore{52.7}{0.3} 
& \appuscore{56.1}{0.4} 
& \appbscore{56.3}{0.1} \\

RefCOCO+ (testA) 
& 69.5 
& \appscore{55.7}{0.5} 
& \appuscore{59.2}{0.2} 
& \appbscore{61.9}{0.3} 
& \appscore{54.3}{0.3} 
& \appuscore{58.0}{0.3} 
& \appbscore{61.9}{0.3} 
& \appscore{50.0}{0.1} 
& \appuscore{52.6}{0.5} 
& \appbscore{53.6}{0.3} \\

RefCOCO+ (testB) 
& 61.4 
& \appscore{49.3}{0.6} 
& \appuscore{51.5}{0.5} 
& \appbscore{54.7}{0.1} 
& \appscore{47.4}{0.6} 
& \appuscore{50.1}{0.4} 
& \appbscore{54.8}{0.2} 
& \appscore{44.4}{0.5} 
& \appuscore{47.2}{0.0} 
& \appbscore{49.0}{0.4} \\

RefCOCO+ (val) 
& 65.3 
& \appscore{52.8}{0.4} 
& \appuscore{55.9}{0.2} 
& \appbscore{58.7}{0.3} 
& \appscore{51.3}{0.3} 
& \appuscore{54.5}{0.1} 
& \appbscore{58.7}{0.3} 
& \appscore{47.4}{0.2} 
& \appuscore{50.3}{0.4} 
& \appbscore{51.5}{0.1} \\

RefCOCO-g (test) 
& 65.4 
& \appscore{51.3}{0.3} 
& \appuscore{55.0}{0.1} 
& \appbscore{57.8}{0.1} 
& \appscore{50.1}{0.4} 
& \appuscore{53.7}{0.3} 
& \appbscore{57.9}{0.1} 
& \appscore{46.8}{0.2} 
& \appuscore{50.4}{0.0} 
& \appbscore{51.5}{0.1} \\

RefCOCO-g (val) 
& 64.9 
& \appscore{51.1}{0.6} 
& \appuscore{54.3}{0.3} 
& \appbscore{57.6}{0.1} 
& \appscore{49.8}{0.5} 
& \appuscore{53.3}{0.4} 
& \appbscore{57.7}{0.3} 
& \appscore{46.4}{0.1} 
& \appuscore{49.9}{0.2} 
& \appbscore{51.4}{0.1} \\

\midrule
\textbf{Mean Retention} 
& -- 
& 81.7\% & \underline{86.4\%} & \textbf{90.6\%} 
& 79.6\% & \underline{84.9\%} & \textbf{90.8\%} 
& 74.2\% & \underline{79.0\%} & \textbf{80.5\%} \\
\bottomrule
\end{tabular}
}
}
\end{table*}
\begin{table*}[t]
\centering
\caption{\textbf{Detailed Performance on Resolution-Sensitive Benchmarks.}
Raw scores are reported as the mean over three random seeds, with standard deviation shown in gray. Per-benchmark retention values are omitted for readability and aggregate retention relative to Vanilla PG2 is reported in the final row. Vanilla PG2 is shaded in grey and serves as the uncompressed reference. Best results per budget are \textbf{bolded}, and second-best are \underline{underlined}.}
\label{tab:detailed_res}
{
\setlength{\tabcolsep}{3.5pt}
\renewcommand{\arraystretch}{1.18}
\resizebox{\textwidth}{!}{%
\begin{tabular}{@{}l >{\columncolor[gray]{0.92}}c ccc ccc ccc@{}}
\toprule
& \multicolumn{1}{c}{} 
& \multicolumn{3}{c}{\textbf{256 Visual Tokens}} 
& \multicolumn{3}{c}{\textbf{64 Visual Tokens}} 
& \multicolumn{3}{c}{\textbf{16 Visual Tokens}} \\
\cmidrule(lr){3-5} \cmidrule(lr){6-8} \cmidrule(lr){9-11}
\appbenchhead & \appmethodhead{PG2}
& \appmethodhead{M$^3$} & \appmethodhead{MQT} & \apparcelhead
& \appmethodhead{M$^3$} & \appmethodhead{MQT} & \apparcelhead
& \appmethodhead{M$^3$} & \appmethodhead{MQT} & \apparcelhead \\
\midrule

ChartQA (aug) 
& 72.5 
& \appscore{64.1}{0.6} 
& \appuscore{66.0}{0.5} 
& \appbscore{66.2}{0.8} 
& \appscore{63.0}{1.3} 
& \appuscore{64.8}{0.2} 
& \appbscore{66.3}{0.2} 
& \appbscore{59.9}{0.8} 
& \appscore{58.2}{0.9} 
& \appuscore{58.8}{0.9} \\

ChartQA (human) 
& 40.7 
& \appscore{33.8}{0.3} 
& \appuscore{35.6}{0.8} 
& \appbscore{37.2}{0.6} 
& \appscore{32.3}{1.0} 
& \appuscore{34.1}{0.9} 
& \appbscore{37.0}{0.3} 
& \appscore{29.7}{0.7} 
& \appuscore{30.7}{0.3} 
& \appbscore{32.0}{1.0} \\

DocVQA (val) 
& 36.6 
& \appscore{30.5}{0.3} 
& \appbscore{33.4}{0.1} 
& \appuscore{32.8}{0.3} 
& \appscore{28.7}{0.1} 
& \appuscore{31.2}{0.5} 
& \appbscore{32.1}{0.4} 
& \appuscore{25.9}{0.2} 
& \appscore{24.4}{0.1} 
& \appbscore{26.1}{0.1} \\

InfoVQA (val) 
& 24.8 
& \appbscore{25.0}{0.2} 
& \appscore{24.2}{0.3} 
& \appuscore{24.4}{0.3} 
& \appscore{23.4}{0.3} 
& \appuscore{23.6}{0.2} 
& \appbscore{23.9}{0.5} 
& \appbscore{22.9}{0.2} 
& \appuscore{22.3}{0.3} 
& \appscore{22.2}{0.2} \\

SciCap 
& 163.5 
& \appscore{159.6}{0.7} 
& \appuscore{163.8}{0.5} 
& \appbscore{164.4}{2.0} 
& \appscore{159.9}{0.7} 
& \appbscore{164.2}{0.2} 
& \appuscore{163.6}{2.7} 
& \appscore{159.1}{0.1} 
& \appbscore{160.4}{0.5} 
& \appuscore{159.7}{2.8} \\

ST-VQA (val) 
& 59.9 
& \appbscore{60.4}{0.2} 
& \appuscore{59.7}{0.3} 
& \appscore{59.1}{0.1} 
& \appbscore{59.3}{0.1} 
& \appscore{58.1}{0.5} 
& \appuscore{58.9}{0.1} 
& \appbscore{56.8}{0.3} 
& \appuscore{54.9}{0.1} 
& \appscore{54.5}{0.4} \\

TextCaps 
& 124.6 
& \appscore{123.2}{1.0} 
& \appuscore{125.0}{0.2} 
& \appbscore{125.8}{0.9} 
& \appscore{121.9}{0.5} 
& \appuscore{123.4}{0.8} 
& \appbscore{123.8}{0.5} 
& \appbscore{115.8}{0.5} 
& \appuscore{114.1}{0.3} 
& \appscore{113.0}{0.7} \\

TextVQA (val) 
& 57.7 
& \appbscore{57.9}{0.2} 
& \appuscore{57.5}{0.1} 
& \appscore{57.2}{0.2} 
& \appbscore{56.6}{0.1} 
& \appscore{56.5}{0.3} 
& \appuscore{56.6}{0.1} 
& \appbscore{54.3}{0.5} 
& \appuscore{52.8}{0.3} 
& \appscore{52.1}{0.2} \\

WidgetCap 
& 133.8 
& \appuscore{133.9}{0.2} 
& \appscore{133.5}{1.1} 
& \appbscore{134.4}{0.9} 
& \appuscore{132.9}{0.2} 
& \appscore{131.9}{0.6} 
& \appbscore{133.4}{1.1} 
& \appbscore{130.9}{1.1} 
& \appuscore{129.4}{0.4} 
& \appscore{128.1}{1.3} \\

\midrule
\textbf{Mean Retention} 
& -- 
& 94.8\% & \underline{96.3\%} & \textbf{96.7\%} 
& 92.3\% & \underline{94.1\%} & \textbf{95.7\%} 
& \textbf{88.4\%} & 86.8\% & \underline{87.3\%} \\
\bottomrule
\end{tabular}
}
}
\end{table*}
\begin{table*}[t]
\centering
\begingroup
\providecommand{\appscore}[2]{\ensuremath{#1\,{\scriptstyle\textcolor{gray}{\pm #2}}}}
\providecommand{\appbscore}[2]{\ensuremath{\mathbf{#1}\,{\scriptstyle\textcolor{gray}{\pm #2}}}}
\providecommand{\appuscore}[2]{\underline{\ensuremath{#1\,{\scriptstyle\textcolor{gray}{\pm #2}}}}}
\providecommand{\appmethodhead}[1]{\makecell[c]{\textbf{#1}\\[-2pt]\scriptsize\vphantom{\textbf{\textit{(Ours)}}}}}
\providecommand{\apparcelhead}{\makecell[c]{\textbf{PARCEL}\\[-2pt]\footnotesize\textbf{\textit{(Ours)}}}}
\providecommand{\appbenchhead}{\makecell[l]{\textbf{Benchmark}\\[-2pt]\scriptsize\vphantom{\textbf{\textit{(Ours)}}}}}
\caption{\textbf{Detailed Performance on General Vision-Language Benchmarks.}
Raw scores are reported as the mean over three random seeds, with standard deviation shown in gray. Per-benchmark retention values are omitted for readability and aggregate retention relative to Vanilla PG2 is reported in the final row. Vanilla PG2 is shaded in grey and serves as the uncompressed reference. Best results per budget are \textbf{bolded}, and second-best are \underline{underlined}.}
\label{tab:detailed_general}
{
\setlength{\tabcolsep}{3.5pt}
\renewcommand{\arraystretch}{1.18}
\resizebox{\textwidth}{!}{%
\begin{tabular}{@{}l >{\columncolor[gray]{0.92}}c ccc ccc ccc@{}}
\toprule
& \multicolumn{1}{c}{} 
& \multicolumn{3}{c}{\textbf{256 Visual Tokens}} 
& \multicolumn{3}{c}{\textbf{64 Visual Tokens}} 
& \multicolumn{3}{c}{\textbf{16 Visual Tokens}} \\
\cmidrule(lr){3-5} \cmidrule(lr){6-8} \cmidrule(lr){9-11}
\appbenchhead & \appmethodhead{PG2}
& \appmethodhead{M$^3$} & \appmethodhead{MQT} & \apparcelhead
& \appmethodhead{M$^3$} & \appmethodhead{MQT} & \apparcelhead
& \appmethodhead{M$^3$} & \appmethodhead{MQT} & \apparcelhead \\
\midrule

AI2D 
& 74.4 
& \appscore{71.6}{0.5} 
& \appuscore{72.9}{0.8} 
& \appbscore{74.2}{0.4} 
& \appscore{71.0}{0.5} 
& \appuscore{72.3}{0.6} 
& \appbscore{74.1}{0.2} 
& \appscore{69.7}{0.3} 
& \appuscore{69.8}{0.6} 
& \appbscore{71.4}{0.7} \\

AOKVQA-DA (val) 
& 62.4 
& \appscore{61.6}{0.5} 
& \appbscore{62.6}{0.3} 
& \appuscore{61.7}{0.7} 
& \appbscore{61.7}{0.4} 
& \appbscore{61.7}{0.7} 
& \appuscore{61.2}{0.3} 
& \appbscore{60.3}{0.4} 
& \appbscore{60.3}{0.4} 
& \appbscore{60.3}{0.4} \\

AOKVQA-MC (val) 
& 79.7 
& \appscore{77.6}{0.6} 
& \appbscore{78.6}{0.6} 
& \appuscore{78.2}{0.1} 
& \appuscore{77.4}{0.7} 
& \appbscore{77.6}{0.3} 
& \appscore{75.0}{1.9} 
& \appbscore{74.9}{0.7} 
& \appbscore{74.9}{0.7} 
& \appbscore{74.9}{0.7} \\

COCO-35L (en) 
& 138.1 
& \appscore{138.6}{0.3} 
& \appbscore{139.0}{0.5} 
& \appuscore{138.8}{0.5} 
& \appbscore{138.4}{0.3} 
& \appuscore{138.2}{0.5} 
& \appscore{138.1}{0.5} 
& \appbscore{134.9}{0.2} 
& \appbscore{134.9}{0.2} 
& \appbscore{134.9}{0.2} \\

COCO-Captions 
& 141.1 
& \appuscore{140.2}{0.1} 
& \appbscore{140.5}{0.7} 
& \appbscore{140.5}{0.2} 
& \appuscore{139.6}{0.4} 
& \appbscore{140.5}{0.3} 
& \appscore{138.2}{0.3} 
& \appbscore{135.9}{0.4} 
& \appbscore{135.9}{0.4} 
& \appbscore{135.9}{0.4} \\

CountBenchQA 
& 80.2 
& \appbscore{81.0}{1.1} 
& \appscore{79.7}{2.4} 
& \appuscore{80.1}{0.7} 
& \appuscore{80.3}{1.2} 
& \appbscore{80.5}{0.9} 
& \appscore{77.6}{0.6} 
& \appbscore{78.9}{1.7} 
& \appbscore{78.9}{1.7} 
& \appbscore{78.9}{1.7} \\

GQA 
& 65.6 
& \appbscore{65.1}{0.1} 
& \appscore{64.7}{0.3} 
& \appuscore{65.0}{0.1} 
& \appbscore{64.7}{0.4} 
& \appscore{64.3}{0.5} 
& \appbscore{64.7}{0.1} 
& \appbscore{63.5}{0.4} 
& \appscore{62.2}{0.4} 
& \appuscore{62.6}{0.1} \\

NLVR2 
& 90.8 
& \appuscore{89.9}{0.2} 
& \appscore{89.2}{0.4} 
& \appbscore{90.1}{0.3} 
& \appuscore{89.8}{0.1} 
& \appscore{88.9}{0.3} 
& \appbscore{90.3}{0.5} 
& \appuscore{88.7}{0.0} 
& \appscore{86.8}{0.2} 
& \appbscore{88.8}{0.3} \\

NoCaps 
& 122.2 
& \appuscore{121.9}{0.3} 
& \appscore{121.2}{0.2} 
& \appbscore{122.0}{0.3} 
& \appscore{120.8}{0.3} 
& \appuscore{120.9}{0.2} 
& \appbscore{121.6}{0.6} 
& \appbscore{119.0}{0.3} 
& \appuscore{118.7}{0.2} 
& \appscore{117.6}{0.7} \\

OCR-VQA 
& 72.2 
& \appbscore{72.1}{0.0} 
& \appbscore{72.1}{0.0} 
& \appuscore{72.0}{0.1} 
& \appuscore{71.6}{0.0} 
& \appscore{71.5}{0.0} 
& \appbscore{71.7}{0.0} 
& \appbscore{69.5}{0.0} 
& \appuscore{67.7}{0.1} 
& \appscore{67.3}{0.1} \\

OKVQA 
& 63.4 
& \appscore{61.6}{0.5} 
& \appbscore{62.6}{0.3} 
& \appuscore{61.7}{0.7} 
& \appbscore{61.7}{0.4} 
& \appbscore{61.7}{0.7} 
& \appuscore{61.2}{0.3} 
& \appbscore{60.3}{0.4} 
& \appbscore{60.3}{0.4} 
& \appbscore{60.3}{0.4} \\

RSVQA-hr (test) 
& 92.6 
& \appbscore{92.7}{0.1} 
& \appbscore{92.7}{0.0} 
& \appbscore{92.7}{0.1} 
& \appbscore{92.7}{0.1} 
& \appbscore{92.7}{0.1} 
& \appbscore{92.7}{0.1} 
& \appbscore{92.6}{0.0} 
& \appbscore{92.6}{0.0} 
& \appbscore{92.6}{0.0} \\

RSVQA-hr (test2) 
& 90.6 
& \appbscore{90.8}{0.2} 
& \appscore{90.5}{0.1} 
& \appuscore{90.7}{0.0} 
& \appbscore{90.8}{0.2} 
& \appuscore{90.5}{0.2} 
& \appbscore{90.8}{0.1} 
& \appbscore{90.7}{0.1} 
& \appscore{90.4}{0.1} 
& \appuscore{90.5}{0.1} \\

RSVQA-lr 
& 92.7 
& \appuscore{92.8}{1.0} 
& \appscore{92.5}{0.3} 
& \appbscore{93.0}{0.5} 
& \appscore{93.2}{0.9} 
& \appuscore{93.3}{0.2} 
& \appbscore{93.4}{0.9} 
& \appbscore{92.6}{0.5} 
& \appbscore{92.6}{0.5} 
& \appbscore{92.6}{0.5} \\

Screen2Words 
& 113.5 
& \appscore{112.3}{1.4} 
& \appuscore{112.5}{1.1} 
& \appbscore{113.2}{0.7} 
& \appuscore{112.2}{1.5} 
& \appscore{111.6}{0.5} 
& \appbscore{112.8}{1.0} 
& \appbscore{111.6}{0.7} 
& \appscore{109.7}{0.5} 
& \appuscore{111.4}{0.8} \\

TallyQA (complex) 
& 69.9 
& \appbscore{68.5}{0.0} 
& \appscore{67.5}{0.1} 
& \appuscore{68.1}{0.2} 
& \appuscore{67.4}{0.3} 
& \appscore{66.0}{0.3} 
& \appbscore{67.9}{0.4} 
& \appbscore{65.7}{0.4} 
& \appscore{64.5}{0.6} 
& \appuscore{65.0}{0.3} \\

TallyQA (simple) 
& 81.0 
& \appbscore{80.5}{0.1} 
& \appscore{80.0}{0.2} 
& \appuscore{80.4}{0.0} 
& \appuscore{80.0}{0.0} 
& \appscore{79.6}{0.1} 
& \appbscore{80.4}{0.1} 
& \appbscore{79.3}{0.0} 
& \appscore{78.2}{0.1} 
& \appuscore{78.7}{0.2} \\

ScienceQA 
& 96.5 
& \appuscore{94.7}{1.1} 
& \appbscore{95.7}{0.6} 
& \appbscore{95.7}{0.3} 
& \appuscore{94.3}{1.3} 
& \appbscore{95.7}{0.2} 
& \appscore{94.2}{1.3} 
& \appbscore{94.3}{0.4} 
& \appbscore{94.3}{0.4} 
& \appbscore{94.3}{0.4} \\

VQAv2 (minival) 
& 82.6 
& \appbscore{81.8}{0.2} 
& \appuscore{81.7}{0.2} 
& \appscore{81.6}{0.3} 
& \appscore{81.1}{0.1} 
& \appuscore{81.2}{0.1} 
& \appbscore{81.5}{0.2} 
& \appbscore{79.8}{0.1} 
& \appuscore{79.2}{0.1} 
& \appscore{78.6}{0.1} \\

VizWizVQA (val) 
& 75.9 
& \appbscore{76.0}{0.2} 
& \appuscore{75.6}{0.7} 
& \appbscore{76.0}{0.1} 
& \appuscore{75.6}{0.4} 
& \appscore{75.5}{0.4} 
& \appbscore{75.7}{0.2} 
& \appbscore{75.1}{0.3} 
& \appuscore{74.6}{0.3} 
& \appscore{74.3}{0.3} \\

XM3600 (en) 
& 80.3 
& \appuscore{79.6}{0.2} 
& \appbscore{80.1}{0.3} 
& \appscore{78.5}{0.4} 
& \appbscore{79.7}{0.3} 
& \appuscore{79.3}{0.5} 
& \appscore{78.6}{0.3} 
& \appbscore{79.4}{1.0} 
& \appuscore{77.7}{0.2} 
& \appscore{77.5}{0.1} \\

\midrule
\textbf{Mean Retention} 
& -- 
& 99.2\% & \underline{99.2\%} & \textbf{99.4\%} 
& \underline{98.8\%} & 98.6\% & \textbf{99.2\%} 
& \textbf{97.6\%} & \underline{96.9\%} & 96.8\% \\
\bottomrule
\end{tabular}
}
}
\endgroup
\end{table*}

\subsection{Detailed Results for Benchmarks at High Resolution}
\label{app-sec:detailed-results-high-resolution}
\begin{table*}[t]
\centering
\caption{\textbf{Detailed Performance on High-Resolution \(448\times448\) Benchmarks.}
Raw scores are reported for single-seed high-resolution fine-tuning runs due to computational cost. Per-benchmark retention values are omitted for readability and aggregate retention relative to the \(448\times448\) Vanilla PG2 reference is reported in the final row. Vanilla PG2 is shaded in grey and serves as the uncompressed reference. Best results per budget are \textbf{bolded}, and second-best are \underline{underlined}.}
\label{tab:app_448_all}
{
\setlength{\tabcolsep}{3.0pt}
\renewcommand{\arraystretch}{1.08}
\resizebox{0.9\textwidth}{!}{%
\begin{tabular}{@{}l >{\columncolor[gray]{0.92}}c ccc ccc ccc@{}}
\toprule
& \multicolumn{1}{c}{} 
& \multicolumn{3}{c}{\textbf{1024 Visual Tokens}} 
& \multicolumn{3}{c}{\textbf{256 Visual Tokens}} 
& \multicolumn{3}{c}{\textbf{64 Visual Tokens}} \\
\cmidrule(lr){3-5} \cmidrule(lr){6-8} \cmidrule(lr){9-11}
\textbf{Benchmark} & \textbf{PG2} 
& \textbf{M$^3$} & \textbf{MQT} & \textbf{\ours}
& \textbf{M$^3$} & \textbf{MQT} & \textbf{\ours}
& \textbf{M$^3$} & \textbf{MQT} & \textbf{\ours} \\
\midrule

\multicolumn{11}{@{}l}{\textit{Resolution-Sensitive Benchmarks}} \\
\midrule
ChartQA (aug) 
& 88.3 
& \(\underline{85.0}\) & \(83.5\) & \(\mathbf{85.8}\)
& \(83.4\) & \(\underline{84.3}\) & \(\mathbf{85.6}\)
& \(\underline{82.3}\) & \(81.8\) & \(\mathbf{82.7}\) \\

ChartQA (human) 
& 53.2 
& \(\underline{48.2}\) & \(48.1\) & \(\mathbf{50.4}\)
& \(48.2\) & \(\underline{48.7}\) & \(\mathbf{50.2}\)
& \(45.6\) & \(\underline{46.2}\) & \(\mathbf{47.5}\) \\

DocVQA (val) 
& 69.8 
& \(\underline{62.9}\) & \(61.8\) & \(\mathbf{63.7}\)
& \(57.8\) & \(\underline{60.4}\) & \(\mathbf{63.7}\)
& \(\underline{50.3}\) & \(49.4\) & \(\mathbf{52.5}\) \\

InfoVQA (val) 
& 35.3 
& \(\mathbf{35.1}\) & \(31.8\) & \(\underline{34.0}\)
& \(\underline{33.2}\) & \(30.9\) & \(\mathbf{34.2}\)
& \(\underline{29.5}\) & \(27.5\) & \(\mathbf{29.5}\) \\

SciCap 
& 182.3 
& \(177.9\) & \(\underline{179.6}\) & \(\mathbf{183.5}\)
& \(177.2\) & \(\underline{181.1}\) & \(\mathbf{183.4}\)
& \(177.2\) & \(\underline{181.2}\) & \(\mathbf{181.8}\) \\

ST-VQA 
& 78.6 
& \(\mathbf{79.9}\) & \(77.0\) & \(\underline{78.8}\)
& \(\underline{78.9}\) & \(77.9\) & \(\mathbf{78.9}\)
& \(\underline{76.7}\) & \(75.7\) & \(\mathbf{76.9}\) \\

TextCaps 
& 146.2 
& \(145.7\) & \(\underline{146.2}\) & \(\mathbf{149.1}\)
& \(146.1\) & \(\underline{146.7}\) & \(\mathbf{149.2}\)
& \(\underline{144.1}\) & \(142.1\) & \(\mathbf{145.7}\) \\

TextVQA 
& 72.8 
& \(\mathbf{73.2}\) & \(72.0\) & \(\underline{72.6}\)
& \(\mathbf{72.9}\) & \(71.9\) & \(\underline{72.7}\)
& \(\mathbf{71.0}\) & \(\underline{70.0}\) & \(69.9\) \\

WidgetCap 
& 148.8 
& \(\underline{147.1}\) & \(146.1\) & \(\mathbf{147.9}\)
& \(146.1\) & \(\underline{147.4}\) & \(\mathbf{148.1}\)
& \(144.2\) & \(\mathbf{145.6}\) & \(\underline{145.1}\) \\

\midrule
\multicolumn{11}{@{}l}{\textit{Image Segmentation (RefCOCO)}} \\
\midrule
RefCOCO (testA) 
& 76.4 
& \(67.1\) & \(\underline{67.8}\) & \(\mathbf{72.7}\)
& \(66.8\) & \(\underline{68.6}\) & \(\mathbf{72.8}\)
& \(64.8\) & \(\underline{67.2}\) & \(\mathbf{69.2}\) \\

RefCOCO (testB) 
& 71.5 
& \(62.1\) & \(\underline{63.4}\) & \(\mathbf{67.4}\)
& \(62.1\) & \(\underline{64.3}\) & \(\mathbf{67.6}\)
& \(60.6\) & \(\underline{63.4}\) & \(\mathbf{64.7}\) \\

RefCOCO (val) 
& 69.6 
& \(64.9\) & \(\underline{65.7}\) & \(\mathbf{69.9}\)
& \(64.6\) & \(\underline{66.5}\) & \(\mathbf{69.9}\)
& \(62.9\) & \(\underline{64.9}\) & \(\mathbf{66.8}\) \\

RefCOCO+ (testA) 
& 74.0 
& \(\underline{64.2}\) & \(64.1\) & \(\mathbf{69.8}\)
& \(63.3\) & \(\underline{65.3}\) & \(\mathbf{69.5}\)
& \(60.7\) & \(\underline{62.8}\) & \(\mathbf{65.5}\) \\

RefCOCO+ (testB) 
& 65.0 
& \(56.0\) & \(\underline{56.2}\) & \(\mathbf{61.1}\)
& \(55.3\) & \(\underline{56.4}\) & \(\mathbf{61.0}\)
& \(53.8\) & \(\underline{54.4}\) & \(\mathbf{58.0}\) \\

RefCOCO+ (val) 
& 69.8 
& \(\underline{60.4}\) & \(59.9\) & \(\mathbf{65.6}\)
& \(59.7\) & \(\underline{60.9}\) & \(\mathbf{65.8}\)
& \(58.0\) & \(\underline{58.7}\) & \(\mathbf{62.4}\) \\

RefCOCO-g (test) 
& 70.0 
& \(59.1\) & \(\underline{60.1}\) & \(\mathbf{64.8}\)
& \(58.6\) & \(\underline{61.1}\) & \(\mathbf{64.7}\)
& \(56.7\) & \(\underline{59.1}\) & \(\mathbf{61.7}\) \\

RefCOCO-g (val) 
& 69.6 
& \(59.1\) & \(\underline{59.3}\) & \(\mathbf{64.7}\)
& \(58.6\) & \(\underline{60.0}\) & \(\mathbf{64.4}\)
& \(56.5\) & \(\underline{58.0}\) & \(\mathbf{61.1}\) \\

\midrule
\multicolumn{11}{@{}l}{\textit{General Vision-Language Understanding}} \\
\midrule
AI2D 
& 75.1 
& \(\underline{74.9}\) & \(73.7\) & \(\mathbf{75.5}\)
& \(\underline{75.5}\) & \(74.1\) & \(\mathbf{75.8}\)
& \(\mathbf{75.0}\) & \(73.0\) & \(\underline{74.6}\) \\

AOKVQA-DA 
& 65.2 
& \(64.6\) & \(\underline{64.7}\) & \(\mathbf{65.5}\)
& \(64.9\) & \(\underline{65.0}\) & \(\mathbf{65.5}\)
& \(63.3\) & \(\underline{63.5}\) & \(\mathbf{64.7}\) \\

AOKVQA-MC 
& 80.8 
& \(80.1\) & \(\underline{80.8}\) & \(\mathbf{81.7}\)
& \(80.2\) & \(\mathbf{81.3}\) & \(\underline{81.2}\)
& \(79.5\) & \(\underline{79.9}\) & \(\mathbf{81.0}\) \\

COCO-35L 
& 140.4 
& \(\mathbf{141.4}\) & \(140.2\) & \(\underline{141.1}\)
& \(\mathbf{141.9}\) & \(140.8\) & \(\underline{141.0}\)
& \(\underline{141.2}\) & \(139.3\) & \(\mathbf{141.3}\) \\

COCO-Captions 
& 142.1 
& \(142.2\) & \(\underline{142.3}\) & \(\mathbf{143.5}\)
& \(\underline{142.1}\) & \(\underline{142.1}\) & \(\mathbf{142.9}\)
& \(\underline{141.9}\) & \(141.4\) & \(\mathbf{142.0}\) \\

GQA 
& 67.7 
& \(\underline{67.3}\) & \(66.0\) & \(\mathbf{67.4}\)
& \(\mathbf{67.3}\) & \(65.9\) & \(\underline{67.0}\)
& \(\mathbf{66.4}\) & \(65.6\) & \(\underline{66.2}\) \\

NLVR2 
& 91.1 
& \(\underline{89.8}\) & \(88.4\) & \(\mathbf{89.8}\)
& \(\mathbf{90.0}\) & \(89.2\) & \(\underline{89.8}\)
& \(\mathbf{89.8}\) & \(88.5\) & \(\underline{89.3}\) \\

NoCaps 
& 123.5 
& \(\mathbf{121.8}\) & \(121.1\) & \(\underline{121.5}\)
& \(\underline{122.4}\) & \(\mathbf{122.5}\) & \(121.7\)
& \(121.1\) & \(\mathbf{121.6}\) & \(\underline{121.5}\) \\

OCR-VQA 
& 74.9 
& \(\underline{74.7}\) & \(74.4\) & \(\mathbf{74.9}\)
& \(\underline{74.7}\) & \(74.7\) & \(\mathbf{74.8}\)
& \(\mathbf{74.2}\) & \(73.8\) & \(\underline{74.0}\) \\

OKVQA 
& 63.7 
& \(\mathbf{63.0}\) & \(62.0\) & \(\underline{63.0}\)
& \(\underline{62.6}\) & \(62.1\) & \(\mathbf{63.0}\)
& \(61.5\) & \(\underline{61.7}\) & \(\mathbf{62.5}\) \\

RSVQA-hr 
& 92.9 
& \(\underline{92.8}\) & \(92.8\) & \(\mathbf{92.8}\)
& \(\mathbf{92.9}\) & \(92.8\) & \(\underline{92.9}\)
& \(\mathbf{92.9}\) & \(\underline{92.8}\) & \(92.8\) \\

RSVQA-hr (test2) 
& 90.8 
& \(\underline{90.7}\) & \(\mathbf{90.8}\) & \(90.6\)
& \(\underline{90.7}\) & \(\mathbf{90.8}\) & \(90.6\)
& \(\mathbf{90.8}\) & \(\mathbf{90.8}\) & \(\underline{90.6}\) \\

ScienceQA 
& 96.0 
& \(\mathbf{95.7}\) & \(95.5\) & \(\underline{95.6}\)
& \(\mathbf{95.8}\) & \(95.5\) & \(\underline{95.7}\)
& \(\mathbf{95.4}\) & \(94.8\) & \(\underline{95.2}\) \\

Screen2Words 
& 114.2 
& \(\underline{115.1}\) & \(113.1\) & \(\mathbf{115.7}\)
& \(\underline{115.6}\) & \(114.5\) & \(\mathbf{116.7}\)
& \(\underline{115.1}\) & \(113.8\) & \(\mathbf{116.1}\) \\

VQAv2 (val) 
& 84.2 
& \(\underline{83.4}\) & \(82.9\) & \(\mathbf{84.0}\)
& \(\underline{83.3}\) & \(82.9\) & \(\mathbf{84.0}\)
& \(\underline{82.7}\) & \(82.4\) & \(\mathbf{83.4}\) \\

VizWizVQA 
& 77.1 
& \(76.2\) & \(\underline{76.8}\) & \(\mathbf{77.0}\)
& \(\underline{76.8}\) & \(76.5\) & \(\mathbf{77.2}\)
& \(\underline{76.6}\) & \(76.0\) & \(\mathbf{76.7}\) \\

XM3600 
& 80.0 
& \(\underline{81.0}\) & \(80.1\) & \(\mathbf{81.0}\)
& \(\underline{80.7}\) & \(79.8\) & \(\mathbf{81.2}\)
& \(\underline{79.6}\) & \(78.2\) & \(\mathbf{80.8}\) \\

\midrule
\textbf{Mean Retention} 
& -- 
& \underline{96.0\%} & 95.4\% & \textbf{98.2\%}
& 95.4\% & \underline{95.8\%} & \textbf{98.2\%}
& 93.4\% & \underline{93.5\%} & \textbf{95.4\%} \\
\bottomrule
\end{tabular}
}
}
\end{table*}
In this section, we present additional results after pretraining \ours and the baselines~\cite{MQT,M3} with the PaliGemma-2 high-resolution Stage-2 recipe at \(448\times448\) resolution~\cite{PG,PG2}.
Due to the substantially higher cost of high-resolution pretraining and evaluation, these results are reported from a single seed and exclude video benchmarks.
\paragraph{High-Resolution Evaluation.}
Table~\ref{tab:app_448_all} reports detailed results for high-resolution \(448\times448\) PaliGemma-2 evaluations across 1024 (full budget), 256, and 64 visual-token budgets.
We therefore use this analysis as a high-resolution stress test of compression behavior rather than as a replacement for the three-seed default-resolution evaluation.

Overall, \ours retains the strongest aggregate performance across all high-resolution token budgets.
At 1024 visual tokens, \ours reaches \(98.2\%\) mean retention relative to the \(448\times448\) Vanilla PG2 reference, compared to \(96.0\%\) for \(\text{M}^3\) and \(95.4\%\) for MQT.
At 256 tokens, \ours again achieves \(98.2\%\) mean retention, outperforming MQT (\(95.8\%\)) and \(\text{M}^3\) (\(95.4\%\)).
Even under the more constrained 64-token budget, \ours remains the strongest compressed model with \(95.4\%\) mean retention, compared to \(93.5\%\) for MQT and \(93.4\%\) for \(\text{M}^3\).

The detailed breakdown shows that the advantage of \ours is most pronounced on the task families targeted by our design.
On image segmentation benchmarks, \ours consistently improves over both baselines across RefCOCO, RefCOCO+, and RefCOCO-g splits, supporting the role of explicit spatial anchors for preserving layout-sensitive evidence.
On resolution-sensitive tasks such as ChartQA, DocVQA, InfoVQA, and TextCaps, \ours also achieves strong retention, indicating that the hybrid decomposition remains effective when visual inputs are processed at higher spatial resolution.
For general multimodal benchmarks, the gaps are smaller because many tasks already saturate near the Vanilla PG2 reference, but \ours still provides the best aggregate trade-off.
These results suggest that the benefits of spectral partitioning and pool-conditioned query resampling are not limited to the default-resolution setting, but continue to hold under high-resolution visual encoding.
\section{FLOP and KV-Cache Calculations}
\label{app-sec:flop-calculations}

This section details the theoretical FLOP and KV-cache calculations used for the efficiency analysis in Figure~\ref{fig:retention_flops_memory}.
We estimate inference-time compute for the visual encoder, visual connector, cross-modal projection, language decoder, and output head.
Following standard transformer FLOP accounting, one multiply-add is counted as two FLOPs.
Lower-order operations such as normalization, activation functions, positional operations, and softmax normalization are omitted.
Finally, our calculations below also take the text prefix tokens into account instead of omitting them for a realistic estimation of the true operating costs of both the baseline PaliGemma-2 and \ours.

\paragraph{Architectural constants.}
We use the PaliGemma-2 3B configuration, which consists of a SigLIP-So400M vision encoder and a Gemma-2 2B language decoder.
We note that the vision encoder has \(L_v=27\) transformer layers, hidden dimension \(D_v=1152\), and MLP dimension \(M_v=4304\).
Furthermore, the language decoder has \(L_l=26\) layers, hidden dimension \(D_l=2304\), MLP dimension \(M_l=9216\), query heads \(H_q=8\), key-value heads \(H_{kv}=4\), head dimension \(d_h=256\), and vocabulary size \(V=257152\).

\paragraph{Token-count notation.}
For a single image, \(T=1\) and for the video setting, \(T=16\).
At the default \(224\times224\) resolution, the SigLIP encoder produces \(N_v=256\) visual tokens per frame.
Let \(B\) denote the compressed visual-token budget per frame.
For \ours, the budget is decomposed into \(N_p\) spatial anchor tokens and \(N_q\) query tokens:

\begin{equation}
    B = N_{\mathrm{vis}} = N_p + N_q .
\end{equation}

The routing rule as defined in Section \ref{sec:budget-aware-routing-method} is:

\begin{equation}
    (N_p,N_q)
    =
    \begin{cases}
        (16, B-16), & 16 \leq B < 64, \\
        (64, B-64), & 64 \leq B \leq 256.
    \end{cases}
\end{equation}

Then, for \(T\) frames, the number of compressed visual tokens entering the language decoder is:

\begin{equation}
    N_{\mathrm{img}} = T B .
\end{equation}

Furthermore, let \(N_t\) denote the number of text-prefix tokens.
These then result in the following full prefill sequence length:

\begin{equation}
    N_{\mathrm{tot}} = N_{\mathrm{img}} + N_t .
\end{equation}

For the values reported in Figure~\ref{fig:retention_flops_memory}, we use \(N_t=128 +1\) for image inputs and \(N_t=64 + 1\) for 16-frame video inputs, following the official PaliGemma parameters for the image and video evals \cite{PG, PG2}.

\paragraph{Vision encoder FLOPs.}
Each frame is independently encoded by the SigLIP vision encoder.
For one ViT layer, the \(Q,K,V\) projections and output projection cost \(8N_vD_v^2\), the attention matrix products cost \(4N_v^2D_v\), and the two-layer MLP costs \(4N_vD_vM_v\).
Accordingly, the per-frame vision-encoder cost is:

\begin{equation}
    C_{\mathrm{ViT}}^{\mathrm{frame}}
    =
    L_v
    \left(
        8N_vD_v^2
        +
        4N_v^2D_v
        +
        4N_vD_vM_v
    \right),
\end{equation}

Thus, the total vision-encoder cost is given by:

\begin{equation}
    C_{\mathrm{ViT}} = T C_{\mathrm{ViT}}^{\mathrm{frame}} .
\end{equation}

\paragraph{Visual connector FLOPs.}
For the compact efficiency table, we report the \ours connector cost.
At matched visual-token budgets, \(M^3\), MQT, and \ours share the same dominant ViT and LLM costs; their FLOPs differ only in the relatively small connector terms.
For \ours, the connector consists of Query \(\leftrightarrow\) Pool self-attention followed by Query \(\rightarrow\) ViT cross-attention whenever \(N_q>0\).
When \(N_q=0\), the routing naturally reduces to a spatial-anchor-only representation and the query pathway is inactive.

For \(N_q>0\), the Query \(\leftrightarrow\) Pool self-attention over \(B\) compressed tokens costs:

\begin{equation}
    C_{\mathrm{QP}}
    =
    8BD_v^2
    +
    4B^2D_v .
\end{equation}

The Query \(\rightarrow\) ViT cross-attention uses \(N_q\) query tokens to attend to the \(N_v\) original ViT tokens:

\begin{equation}
    C_{\mathrm{Q\rightarrow V}}
    =
    4(N_q+N_v)D_v^2
    +
    4N_qN_vD_v .
\end{equation}

Furthermore, query-token MLP cost is given by:
\begin{equation}
    C_{\mathrm{Q\text{-}MLP}}
    =
    4N_qD_vM_v .
\end{equation}

Following from these, the total connector cost is thus given by:

\begin{equation}
    C_{\mathrm{conn}}
    =
    T \cdot \mathbbm{1}[N_q>0]
    \left(
        C_{\mathrm{QP}}
        +
        C_{\mathrm{Q\rightarrow V}}
        +
        C_{\mathrm{Q\text{-}MLP}}
    \right),
\end{equation}

where \(\mathbbm{1}[N_q>0]\) indicates that the query pathway is active only when query tokens are allocated.

\paragraph{Cross-modal projection FLOPs.}
After compression, the \(B\) visual tokens per frame are projected from the vision dimension \(D_v\) to the language dimension \(D_l\):

\begin{equation}
    C_{\mathrm{proj}}
    =
    2TBD_vD_l .
\end{equation}

\paragraph{Language decoder FLOPs.}
The visual tokens and text prefix are processed by the Gemma-2 language decoder during prefill.
We use the full prefix-attention cost:

\begin{equation}
    A_{\mathrm{prefill}} = N_{\mathrm{tot}}^2 .
\end{equation}

For one Gemma-2 decoder layer with grouped-query attention, the projection cost is:

\begin{equation}
    C_{\mathrm{GQA\text{-}proj}}
    =
    4N_{\mathrm{tot}}D_l d_h(H_q+H_{kv}),
\end{equation}

and the attention matrix cost is:
\begin{equation}
    C_{\mathrm{GQA\text{-}attn}}
    =
    4A_{\mathrm{prefill}}H_qd_h .
\end{equation}

The gated feed-forward network cost is:

\begin{equation}
    C_{\mathrm{FFN}}
    =
    6N_{\mathrm{tot}}D_lM_l .
\end{equation}
The total language-decoder cost is:

\begin{equation}
    C_{\mathrm{LLM}}
    =
    L_l
    \left(
        C_{\mathrm{GQA\text{-}proj}}
        +
        C_{\mathrm{GQA\text{-}attn}}
        +
        C_{\mathrm{FFN}}
    \right).
\end{equation}

\paragraph{Output head FLOPs.}
For the reported FLOP values, we evaluate vocabulary logits over the text-prefix positions.
Thus, with \(N_{\mathrm{logit}}=N_t\), the output projection cost is:

\begin{equation}
    C_{\mathrm{head}}
    =
    2N_tD_lV .
\end{equation}

\paragraph{Total theoretical FLOPs.}
The total theoretical prefill compute is:

\begin{equation}
    C_{\mathrm{total}}
    =
    C_{\mathrm{ViT}}
    +
    C_{\mathrm{conn}}
    +
    C_{\mathrm{proj}}
    +
    C_{\mathrm{LLM}}
    +
    C_{\mathrm{head}} .
\end{equation}

Substituting the constants above and rounding to one decimal place gives the TFLOP values reported in Figure~\ref{fig:retention_flops_memory}.

\paragraph{KV-cache memory.}
We compute KV-cache memory for the language decoder, which is the dominant autoregressive memory term during generation.
Gemma-2 uses grouped-query attention with \(H_{kv}=4\) key-value heads.
Assuming \texttt{bfloat16} cache storage, each scalar requires 2 bytes.
The number of bytes required to store the key and value cache for one token across all decoder layers is:

\begin{equation}
    B_{\mathrm{token}}
    =
    2
    \times
    2
    \times
    L_l
    \times
    H_{kv}
    \times
    d_h ,
\end{equation}

where the first factor of \(2\) is the number of bytes per \texttt{bfloat16} scalar and the second factor of \(2\) accounts for keys and values.
Substituting the Gemma-2 constants gives:

\begin{equation}
    B_{\mathrm{token}}
    =
    2
    \times
    2
    \times
    26
    \times
    4
    \times
    256
    =
    106{,}496
    \;\text{bytes}.
\end{equation}

Thus, the prefill KV-cache memory in MB is:

\begin{equation}
    M_{\mathrm{KV}}
    =
    \frac{
        N_{\mathrm{tot}} B_{\mathrm{token}}
    }{
        1024^2
    }.
\end{equation}
For image inputs, \(N_{\mathrm{tot}}=B+129\) and for 16-frame video inputs, \(N_{\mathrm{tot}}=16B+65\).
Rounding to the nearest MB gives the KV-cache values reported in Figure~\ref{fig:retention_flops_memory}.
\section{Benchmark Details}
\label{app-sec:benchmark-details}
We follow the broad transfer-evaluation protocol used in PaliGemma and PaliGemma-2~\cite{PG,PG2}, covering video understanding, dense spatial grounding, resolution-sensitive reasoning, captioning, and general multimodal comprehension.
Below, we briefly describe the role of each benchmark in our evaluation suite.

\paragraph{Video understanding.}
ActivityNet-CAP~\cite{ActivityNetCap} evaluates dense video captioning, requiring the model to summarize human activities and events from short video clips.
ActivityNet-QA~\cite{ActivityNetQA} evaluates question answering over video content, stressing temporal understanding and action-level reasoning.
MSRVTT-CAP~\cite{MSRVTT-Cap} measures video captioning on diverse web videos, testing whether compressed visual tokens preserve enough temporal and semantic context for generation.
MSRVTT-QA~\cite{MSVD1} evaluates open-ended question answering on MSRVTT videos.
MSVD-QA~\cite{MSVD1,MSVD2} similarly targets video question answering, with a focus on short clips and object/action recognition.
VATEX~\cite{VATEX} is a multilingual video captioning benchmark built around human-annotated video descriptions. Since the official test set is not publicly available and we observe high validation-set variance, we report VATEX only in Section~\ref{app-sec:additional-results}.
Overall, for the video benchmarks, there were slight changes in data splits with respect to the official PaliGemma works \cite{PG, PG2} due to data wipeouts associated with these datasets.

\paragraph{Dense spatial grounding and segmentation.}
The RefCOCO suite evaluates referring expression segmentation, where the model must localize the image region described by a natural-language expression.
RefCOCO and RefCOCO+~\cite{RefCOCO1,RefCOCO2} differ in the style of referring expressions, with RefCOCO+ reducing reliance on absolute location words and therefore requiring stronger visual grounding.
RefCOCO-g~\cite{RefCOCOG} contains longer and more descriptive referring expressions, making it a stronger test of language-conditioned dense localization.
Together, these splits directly probe whether visual-token compression preserves explicit spatial layout and fine-grained object boundaries.

\paragraph{Resolution-sensitive document, chart, OCR, and screen tasks.}
ChartQA~\cite{ChartQA} evaluates question answering over charts, with separate augmented and human-written splits.
DocVQA~\cite{DocVQA} evaluates visual question answering over document images, stressing text recognition, layout understanding, and fine-grained evidence retrieval.
InfoVQA~\cite{InfoVQA} extends document VQA to infographic-style inputs, where information is often distributed across text, icons, tables, and visual layouts.
ST-VQA~\cite{ST-VQA} and TextVQA~\cite{TextVQA} evaluate scene-text understanding, requiring the model to read and reason over text embedded in natural images.
OCR-VQA~\cite{OCR-VQA} focuses on recognizing and reasoning over text in  images.
TextCaps~\cite{TextCaps} evaluates caption generation that must incorporate scene text, testing whether the model can preserve text-sensitive visual evidence under compression.
WidgetCap~\cite{WidgetCap} requires captioning a specific user-interface element, making it sensitive to localized UI structure and fine-grained visual details.
Screen2Words~\cite{Screen2Words} evaluates mobile-screen summarization, requiring the model to produce a concise natural-language description of an interface screen.
SciCap~\cite{hsu2021scicap} evaluates scientific figure captioning, where the model must describe structured visual content such as plots, diagrams, and scientific imagery.

\paragraph{General visual question answering and reasoning.}
VQAv2~\cite{VQAv2} is a broad visual question answering benchmark over natural images and serves as a general-purpose VQA test.
GQA~\cite{GQA} evaluates compositional visual reasoning over scene graphs, stressing object relations and structured reasoning.
xGQA~\cite{xGQA} extends GQA to multilingual settings, testing whether visual reasoning transfers across languages.
OKVQA~\cite{OKVQA} requires outside knowledge in addition to visual understanding.
AOKVQA~\cite{AOKVQA} further targets knowledge-intensive visual question answering and we evaluate both direct-answer (AOKVQA-DA) and multiple-choice (AOKVQA-MC) variants.
AI2D~\cite{AI2D} evaluates diagram understanding on science-style illustrations, requiring the model to interpret arrows, labels, and spatial relations.
ScienceQA~\cite{ScienceQA} evaluates multimodal science question answering, combining visual interpretation with textual and commonsense reasoning.
TallyQA~\cite{TallyQA} targets counting-based visual question answering, with simple and complex splits that differ in the difficulty of counting and relational reasoning.
CountBenchQA~\cite{PG} similarly stresses counting and quantity-sensitive reasoning, though with improved and corrected annotations over TallyQA as described in PaliGemma \cite{PG}.
NLVR2~\cite{NLVR2} evaluates reasoning over paired images and a natural-language statement, requiring the model to jointly inspect multiple visual inputs.
MARVL-5~\cite{MARVL5} extends this style of multi-image reasoning to multilingual and culturally diverse settings.
VizWizVQA~\cite{VizWizVQA} evaluates VQA on images captured by blind or low-vision users, which often contain blur, occlusion, unusual framing, or low visual quality.
RSVQA~\cite{RSVQA} evaluates visual question answering over remote-sensing imagery, with low-resolution and high-resolution subsets that stress geospatial interpretation at different image scales.

\paragraph{Captioning and multilingual image understanding.}
COCO-CAP~\cite{MSCOCO, chen2015microsoft} evaluates standard image captioning on MSCOCO-style natural images.
NoCaps~\cite{NoCaps} evaluates captioning on images containing novel objects beyond the standard COCO object categories, testing open-vocabulary generalization.
COCO-35L~\cite{COCO35L} evaluates multilingual image captioning using COCO captions translated across multiple languages, including an English split and multilingual averages.
XM3600~\cite{COCO35L} evaluates cross-lingual image captioning over a diverse multilingual captioning set, further testing whether compressed visual representations remain useful across language settings.
\section{Implementation Details}
\label{app-sec:implementation-details}
\paragraph{Budget sampling and routing.}
For \ours and all baselines, we follow the architectural choices and budget-sampling strategies of the corresponding elastic compression methods as closely as possible.
For MQT, the visual-token budget is entirely allocated to query tokens.
During training, we sample an even query-token budget from \(\{2,4,\ldots,256\}\) for the default \(224\times224\) setting.
For \ours, the sampled visual-token budget is decomposed into spatial anchor tokens and query tokens according to the routing strategy in Section~\ref{sec:budget-aware-routing-method}.
At \(224\times224\), the SigLIP visual grid contains \(16\times16=256\) tokens.
We use two spatial anchor resolutions: a \(4\times4\) anchor grid with \(N_p=16\) tokens, obtained by \(4\times4\) average pooling, and an \(8\times8\) anchor grid with \(N_p=64\) tokens, obtained by \(2\times2\) average pooling.
We sample an even total budget \(B\in\{16,18,\ldots,256\}\).
If \(16\leq B<64\), we use the \(N_p=16\) anchor grid and allocate \(N_q=B-16\) query tokens.
If \(64\leq B\leq256\), we use the \(N_p=64\) anchor grid and allocate \(N_q=B-64\) query tokens.
Thus, query tokens are used only to fill the remaining budget not explicitly covered by the spatial anchor grid.
For M$^3$, we follow the square-only spatial pooling strategy of the original method~\cite{M3}.
This gives supported token budgets \(\{4,16,64,256\}\) in the \(224\times224\) setting.
In the main experiments, we report the budgets shared with the other methods, namely \(16\), \(64\), and \(256\) visual tokens.

\paragraph{Connector architecture.}
For MQT, we use a single query-to-visual cross-attention block following the official MQT design~\cite{MQT}.
For \ours, we use exactly one Query \(\leftrightarrow\) Pool self-attention block followed by one Query \(\rightarrow\) ViT cross-attention block.
No connector attention block is repeated.
All connector attention blocks operate at the visual hidden width \(D_v=1152\) and use 12 attention heads.
The query tokens form a single ordered query bank shared across routing regimes.
For example, the first 48 query embeddings are shared between the low-budget regime that fills budgets above the 16-token anchor and the higher-budget regime that starts from the 64-token anchor and adds query tokens.
This shared prefix structure is the mechanism that allows nested dropout to support elastic query truncation across budgets.

\paragraph{Pretraining setup.}
For PaliGemma-2, the base initialization follows the unimodal pretraining of its constituent modules, including contrastive vision-language pretraining for SigLIP~\cite{SigLIP} and autoregressive text-only pretraining for Gemma 2~\cite{Gemma2}.
Starting from these unimodal components, we train the Vanilla PaliGemma-2 model for 100M samples using the Stage-1 pretraining recipe described in PaliGemma-2~\cite{PG2}.
We use 100M samples because the PaliGemma and PaliGemma-2 studies show that many transfer benchmarks begin to saturate at this scale~\cite{PG,PG2}, and because 1B-sample pretraining is computationally prohibitive for our comparison across multiple compression methods.
For this Stage-1 pretraining, we follow the PaliGemma-2 configuration without modifying the learning-rate multipliers, data mixture, pretraining task mixture, or Gemma 2 logit soft-capping~\cite{Gemma2,LogitCap}.
The pretraining mixture includes captioning, grounded captioning~\cite{LocCa}, OCR, VQA~\cite{VQATasks}, detection, and instance segmentation tasks~\cite{DetTasks1,PaLI3}, with data drawn from sources such as WebLI~\cite{PaLI,PaLI-X} and CC3M~\cite{CC3M}.
For full details on the pretraining data, task definitions, and splits, we refer the reader to the PaliGemma~\cite{PG} and PaliGemma-2~\cite{PG2} works.

\paragraph{Training infrastructure and optimizer.}
All models are trained in the open-source \texttt{big\_vision} codebase \cite{BigVision} following the PaliGemma training setup~\cite{PG,PG2}, but using Cloud TPUv4 accelerators \cite{TPU}.
During pretraining, data, model parameters, and optimizer state are sharded across devices using the JAX/GSPMD \cite{JAX, GSPMD} fully-sharded data-parallel strategy adopted by PaliGemma.
We use 256 TPUv4 chips, a global batch size of 8192 and no gradient accumulation for pretraining.
Following PaliGemma-2~\cite{PG2}, we use Adam \cite{Adam, AdamW} with default hyperparameters.
For Stage-1 and Stage-2 pretraining of the PaliGemma-2 3B backbone, we use the default PaliGemma learning rate of \(2\times10^{-5}\) multiplied by \(0.5\), following the PaliGemma-2 scaling rule.
We also follow PaliGemma-2 in applying Gemma-2 logit soft-capping \cite{Gemma2, LogitCap} during Stage-1 and Stage-2 pretraining, but not during transfer tuning.

\paragraph{Intermediary pretraining of compressed models.}
After Stage-1 pretraining, we integrate the method-specific connector components for MQT, M$^3$, \ours, and the ablations.
Weights shared with PaliGemma-2, including the vision encoder, language decoder, and cross-modal projection, are initialized from the Stage-1 model, while newly introduced method-specific connector parameters are randomly initialized.
During intermediary pretraining, the SigLIP vision encoder, Gemma-2 language decoder, cross-modal projection, and connector parameters are all trainable, following the fully trainable PaliGemma-2 setup.
The learning-rate schedule for newly introduced connector parameters follows the schedule used for the PaliGemma cross-modal projection.
We then perform an additional 100M-sample intermediary pretraining stage for each compressed model with the full method-specific architecture active.
During this stage, nested dropout is enabled for both MQT and \ours.
Because the newly added connector components must be learned from scratch, we restart the learning-rate schedule rather than resuming the Stage-1 schedule.
For a fair comparison against the uncompressed reference, we also continue Vanilla PaliGemma-2 training for an additional 100M samples.

\paragraph{Transfer tuning and evaluation.}
After pretraining, we perform transfer tuning on the benchmarks described in Appendix~\ref{app-sec:benchmark-details}.
Budget sampling remains active during transfer tuning for all elastic compression methods, using the same budget ranges and random sampling strategy as in pretraining.
For video transfer tasks, visual-token budgets are applied per frame, and frame sampling follows the PaliGemma/PaliGemma-2 evaluation setup~\cite{PG,PG2}.
We do not perform additional hyperparameter tuning for the proposed method or baselines.
For model selection, we follow the PaliGemma and PaliGemma-2 transfer protocols and rely on the corresponding evaluation or validation splits whenever available.
The exact transfer-tuning hyperparameters are benchmark-dependent, so we follow the corresponding PaliGemma and PaliGemma-2 transfer recipes.
All three-seed experiments use the same set of randomly sampled seeds across methods, sampled from the range \([10{,}000,100{,}000]\).

\paragraph{High-resolution \(448\times448\) setting.}
For the high-resolution experiments, we follow the Stage-2 high-resolution pretraining strategy of PaliGemma-2~\cite{PG2}.
This stage uses \(448\times448\) image inputs and is run for 10M samples.
At this resolution, the SigLIP visual grid contains \(32\times32=1024\) visual tokens.
For MQT, we expand the query bank so that budgets up to 1024 visual tokens can be sampled.
For M$^3$, square-only spatial pooling gives supported budgets \(\{4,16,64,256,1024\}\).

For \ours, we extend the default-resolution routing by introducing a third spatial anchor scale.
Specifically, we use anchor sizes \(N_p\in\{16,64,256\}\), corresponding to \(4\times4\), \(8\times8\), and \(16\times16\) anchor grids on the \(32\times32\) source grid.
Equivalently, these are obtained by \(8\times8\), \(4\times4\), and \(2\times2\) average pooling, respectively.
Given a sampled high-resolution budget \(B\), we allocate query tokens to fill the gap above the selected anchor size:
\[
(N_p,N_q)
=
\begin{cases}
(16, B-16), & 16 \leq B < 64, \\
(64, B-64), & 64 \leq B < 256, \\
(256, B-256), & 256 \leq B \leq 1024.
\end{cases}
\]
After the 10M-sample high-resolution pretraining stage, we perform high-resolution transfer tuning and evaluation using the PaliGemma-2 high-resolution protocol.
Due to the substantially higher computational cost of \(448\times448\) pretraining and evaluation, high-resolution benchmark results are reported from a single seed.
\section{Limitations and Social Impact}
\label{app-sec:limitations-social-impact}

\paragraph{Limitations.}
While \ours improves the accuracy--efficiency trade-off for visual-token compression, several limitations remain.
First, our method inherits the limitations of the underlying PaliGemma-2 backbone and its pretraining data.
As with other large vision-language models, the model may reflect biases present in web-scale image-text data, including demographic, cultural, geographic, and linguistic data imbalance.
Second, although token compression reduces inference cost, training and evaluating large multimodal models still requires substantial compute.
This limits accessibility and motivates future work on more compute-efficient training recipes, lightweight ablations, and better low-cost evaluation protocols.

Our method also uses a list of budgets that is specified by the practitioner rather than predicted from the input.
Accordingly, developing an input-adaptive budget predictor that allocates visual tokens dynamically is therefore an important direction for future work.

\paragraph{Social impact.}
The primary goal of this work is to make vision-language models more efficient by reducing the number of visual tokens processed by the language decoder.
Improved visual-token efficiency can lower inference cost, reduce memory usage, and make multimodal systems more accessible in resource-constrained environments.
This may be especially useful for applications involving long videos, high-resolution documents, or multi-image inputs, where uncompressed visual tokens can become prohibitively expensive.

At the same time, efficiency improvements can also make powerful multimodal systems easier to deploy at scale.
As a result, the same concerns that apply to large vision-language models also apply here, including biased predictions, hallucinated visual interpretations, privacy risks in image or video analysis, and potential misuse in automated decision-making.
Our method does not introduce new data sources or novel domain-specific capabilities beyond the underlying model, but it may reduce the computational barrier to using such models.

Overall, we view efficient visual-token compression as a positive step toward more practical and sustainable multimodal models.
By improving the performance retained under strict token budgets, \ours can help reduce compute and memory requirements while preserving strong visual understanding.
Future work on adaptive budget prediction, broader backbone validation, and bias-aware evaluation could further improve both the efficiency and responsible deployment of efficient LVLMs.

\end{document}